\documentclass{article}

\usepackage{ocl_preprint,times}
\usepackage{booktabs}
\usepackage{amsfonts}
\usepackage{amsmath}
\usepackage{amssymb}
\usepackage{graphicx}
\usepackage{caption}
\usepackage{microtype}
\usepackage{xcolor}
\usepackage{float}
\usepackage{algorithm}
\usepackage{algpseudocode}
\usepackage{array}
\usepackage{tabularx}
\usepackage{enumitem}
\usepackage{needspace}
\usepackage{xurl}
\usepackage[hidelinks]{hyperref}

\ifdefined\PAEFLOWSTANDARD
  \newcommand{\revisionsevencolor}{\color{black}}
\else
  \newcommand{\revisionsevencolor}{\color{black}}
\fi
\newcommand{\revisionseven}[1]{{\revisionsevencolor #1}}

\newcommand{\cq}[1]{\textcolor{black} {#1}}
\newcolumntype{Y}{>{\raggedright\arraybackslash}X}

\title{OCL-PDE: A Generative Framework for PDE Inverse Problems with Observation-Complementary Latents}

\author{Ding Yang$^{1}$, Chuqi Chen$^{2}$, Chang Ma$^{1}$, Yang Xiang$^{1,3,4}$\\[4pt]
  {\normalfont\small $^{1}$Department of Mathematics,}\\
  {\normalfont\small The Hong Kong University of Science and Technology, Hong Kong}\\
  {\normalfont\small $^{2}$Department of Mathematics,}\\
  {\normalfont\small University of Michigan, Ann Arbor, USA}\\
  {\normalfont\small $^{3}$Department of Materials,}\\
  {\normalfont\small The Hong Kong University of Science and Technology, Hong Kong}\\
  {\normalfont\small $^{4}$Algorithms of Machine Learning and Autonomous Driving Research Lab,}\\
  {\normalfont\small HKUST Shenzhen-Hong Kong Collaborative Innovation Research Institute,}\\
  {\normalfont\small Shenzhen, China}\\[4pt]
  {\normalfont\small\texttt{dyangar@connect.ust.hk}, \texttt{cchenck@umich.edu}}\\
  {\normalfont\small\texttt{changma@ust.hk}, \texttt{maxiang@ust.hk}}}
\date{}
\hypersetup{
  pdftitle={OCL-PDE: A Generative Framework for PDE Inverse Problems with Observation-Complementary Latents},
  pdfauthor={Ding Yang, Chuqi Chen, Chang Ma, Yang Xiang},
  pdfsubject={Preprint}
}

\begin{document}

\raggedbottom

\maketitle

\begin{abstract}
Partial differential equation (PDE) inverse problems are often ill-posed, making fine-scale details difficult
to recover. We address this problem by introducing a learned
observation-complementary latent representation that preserves
reconstruction-relevant information and is combined with the observation
to reconstruct the unknown field. Building on this representation, we propose OCL-PDE,
a generative framework that encourages the observation to guide
large-scale structure and the latent to supply complementary fine-scale
details. OCL-PDE is built on a physics-aware autoencoder (AE)
and conditional Flow Matching, supporting inverse reconstruction as well as forward PDE prediction. Experiments demonstrate
improved reconstruction accuracy and fine-detail recovery compared
with the evaluated baselines.
\end{abstract}

\section{Introduction}

\begingroup
\cq{Inverse problems governed by partial differential equations (PDEs) seek to
recover unknown coefficients, initial or boundary conditions, or source terms
from indirect observations~\citep{isakov2017inverse}. Given an unknown field
$c$ and observations $y$ produced through a PDE forward process, the task is
to infer $c$ from $y$. The available observations may be coarse, sparse, or
noisy; in time-dependent problems, dissipative evolution may further erase
information about the initial state. Such problems are often ill-posed:
distinct fields can produce similar observations, while local variations or
interfaces may be weakly reflected in the measured response
\citep{isakov2017inverse,stuart2010inverse}. Consequently, an inverse method
may agree with the observations while failing to recover important structures
in the underlying field.
}

These difficulties highlight the need for prior information about the unknown field to guide reconstruction. Generative models offer a data-driven way to learn such information by modeling distributions of physical fields. Existing approaches exploit this idea through joint modeling of coefficient and solution fields or through learned latent representations~\citep{huang2024diffusionpde,zang2025dgenno,wang2026fundiff}. 
However, learning a low-dimensional representation alone does not ensure accurate reconstruction in latent-space inverse methods. The representation must retain information useful for recovering field details, and useful latent variables must be inferred from the available observations. These requirements motivate learning latent representations specifically for observation-conditioned reconstruction.

To address these requirements, we introduce a learned observation-complementary latent representation that preserves reconstruction-relevant information and is combined with observations to recover the unknown field. Observations provide a direct basis for reconstruction, while latent features contribute corrections to field structures that are not adequately recovered from observations alone. This design encourages observations to guide the large-scale structure and the latent to supply complementary information for recovering fine-scale variations.

We implement this idea by proposing a framework, OCL-PDE, which learns and generates \emph{observation-complementary latent} (OCL) representations for inverse reconstruction.
A physics-aware autoencoder learns a forward response predictor and a decoder that combines observations with a compact latent. The decoder includes an observation-based reconstruction pathway and a latent correction pathway.  
After the autoencoder is trained, conditional Flow Matching learns to generate its latent codes from Gaussian noise conditioned on observations. At inference, the decoder combines the generated latent with the observation to reconstruct the unknown field. 
The main contributions of our work are summarized as follows:
\begin{itemize}
  \item 
  We develop a generative framework that combines an autoencoder with Flow Matching to improve fine-scale reconstruction through a direct observation pathway and a compact latent correction pathway, while also enabling forward PDE prediction.
\item We develop a two-stage learning procedure that trains the autoencoder with reconstruction supervision and PDE constraints on predicted responses, then freezes it to learn observation-conditioned latent generation through Flow Matching.
\item We demonstrate strong reconstruction results on both backward Allen–Cahn recovery and
Darcy coefficient inversion under coarse or sparse observations, with particularly substantial
improvements over all evaluated baselines in backward Allen–Cahn recovery.
\item \cq{We conduct controlled ablations which further confirm that the learned latent captures observation-complementary information beyond the observation-only pathway.}
\end{itemize}

\endgroup

\begin{figure}[!ht]
  \centering
  \includegraphics[width=0.9\linewidth]{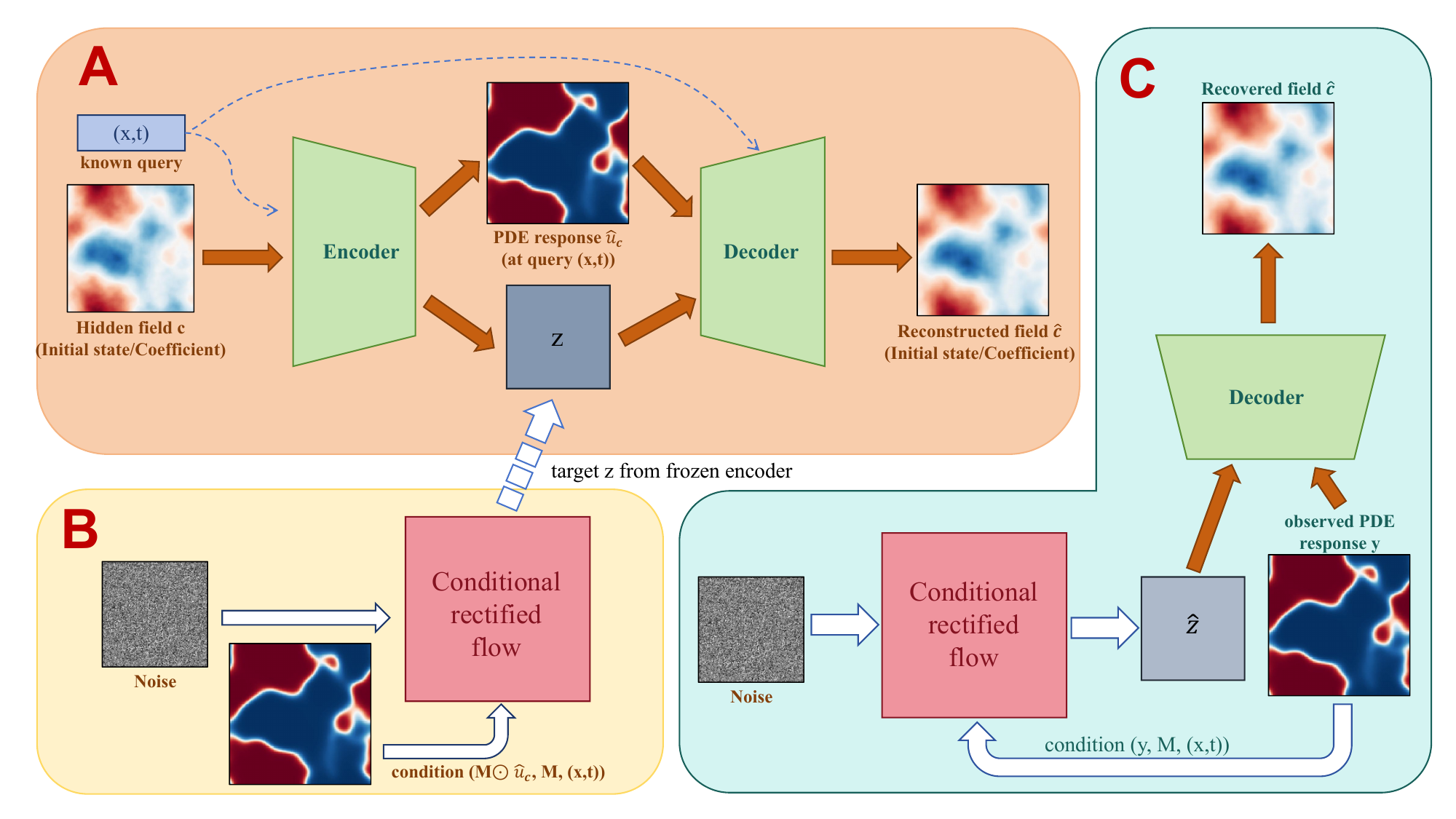}
  \caption{Overview of OCL-PDE. The framework learns
  observation-complementary latents and generates them conditioned on observations
  to recover unknown fields.
  (A) The physics-aware autoencoder predicts the PDE response $\widehat{u}_c$ and encodes
  observation-complementary information in a latent $z$; the decoder combines $\widehat{u}_c$
  and latent $z$ to reconstruct the input field.
  (B) With the autoencoder frozen, the conditional Flow learns to transform
  Gaussian noise into the encoder's latents $z$ guided by the observed response.
  (C) At inference, the Flow generates a latent from noise conditioned on the
  available observation $y$, and the decoder combines the generated latent $\widehat{z}$ and $y$
  to recover the unknown field $c$.}
  \label{fig:two-stage-framework}
\end{figure}

\section{Related Work}

\paragraph{PDE inverse problems.}
Classical inverse formulations recover unknown fields by fitting observations
subject to the governing PDE, often with regularization to stabilize
recovery~\citep{stuart2010inverse}. Numerical methods, including finite element and boundary
element methods, have been developed to solve both forward and inverse PDE
problems~\citep{harrach2021finiteinverse,lesnic2002boundaryinverse}. Learning-based approaches
also combine observations with physical models. PINNs~\citep{raissi2019physics}
use observation data and PDE residuals to infer unknown parameters. The Deep Ritz method~\citep{e2018deepritz} approximates solutions
by minimizing variational energies, while weak adversarial networks
(WANs)~\citep{zang2020weakadversarial,bao2020weakadversarialinverse} recover unknown coefficients and
solutions from measurements using weak PDE constraints.
Neural operators such as DeepONet~\citep{lu2021deeponet} and
FNO~\citep{li2021fourierneuraloperatorparametric} learn mappings between function
spaces. PI-DeepONet~\citep{wang2021learning} and PINO~\citep{li2024pino} further
incorporate PDE constraints into operator learning. These methods do not explicitly model field distributions as generative
priors for inverse reconstruction, potentially limiting fine-detail
recovery when observations are incomplete or degraded.

\paragraph{Generative priors for PDE reconstruction.}
Generative models provide such priors, using learned field distributions
to aid reconstruction from incomplete or degraded observations. Diffusion models,
including DDPM and DDIM~\citep{ho2020ddpm,song2021ddim}, generate samples through
iterative denoising. Diffusion posterior sampling~\citep{chung2023diffusionposterior}
incorporates measurement information into this process for inverse recovery.
DiffusionPDE~\citep{huang2024diffusionpde} learns a joint distribution of
coefficient and solution fields, combining sparse observations and PDE
constraints to guide reconstruction. Related physics-informed diffusion
methods recover high-fidelity flow fields or downscale coarse PDE solutions
\citep{shu2023physicsinformed,lu2024pgdm}, addressing spatial detail missing
from low-fidelity inputs. Conditional Flow Matching has also been applied to
Bayesian inverse problems to generate unknown parameters conditioned on
observations~\citep{sherki2025bayesian}. These formulations generate fields or
their physical parameterizations, without explicitly learning a latent
representation dedicated to complementing observations for fine-detail recovery.

\paragraph{Learned latent representations for PDE reconstruction.}
An alternative is to learn compact field representations and perform inference
in latent space, with VAEs providing a standard generative
formulation~\textcolor{black}{\citep{kingma2014autoencoding}}. FunDiff~\citep{wang2026fundiff}
combines a function autoencoder with latent diffusion to reconstruct and
generate continuous fields under physical constraints.
DGenNO~\citep{zang2025dgenno} jointly represents coefficient and solution fields
through shared latent variables and uses physics constraints to learn from
unlabeled inputs. IGNO~\citep{bao2025igno} uses low-dimensional coefficient
representations and performs inversion by optimizing latent variables against
observations. These methods learn latent representations of physical
fields. Our framework gives the latent $z$ a more specific role:
to preserve reconstruction-relevant information, particularly fine-scale details.

\Needspace{8\baselineskip}
\section{Methodology}
\label{sec:method}

Let $c\in\mathcal C$ denote the unknown field, such as a
coefficient or initial state. $y$ is the possibly noisy observation:
\begin{equation}
  y=\mathcal F(c)+\varepsilon,
  \label{eq:observation}
\end{equation}
Here, $\mathcal F:\mathcal C\rightarrow\mathcal Y$ combines the governing PDE solution operator with the measurement process; $\varepsilon$ is task-specific measurement noise and is zero for clean observations. The goal is to recover $c$ from $y$.

\textcolor{black}{Our OCL-PDE framework consists of an encoder $E_{\theta_E}$, a decoder $D_{\theta_D}$, and a conditional Flow Matching model $v_{\theta_F}$. The encoder $E_{\theta_E}$ predicts the forward response $\widehat y$ and an observation-complementary latent $z$. Given an observation $y$, the Flow $v_{\theta_F}$ obtain the corresponding $z$ conditioned on $y$. The decoder $D_{\theta_D}$ combines $y$ with $z$ to reconstruct $c$. This process is illustrated in Figure~\ref{fig:two-stage-framework}.}
\subsection{physics-aware autoencoder and Observation-complementary latent}
\label{sec:pae}
\textcolor{black}{In OCL-PDE, the physics-aware autoencoder predicts the forward response and encodes the input field into an observation-complementary latent, both of which its decoder uses to reconstruct the unknown field, as shown in Figure~\ref{fig:two-stage-framework}A.}

\textcolor{black}{The encoder has a shared stem, an response branch that predicts $\widehat y$, and a latent branch that encodes the complete unknown field $c$:}
\begin{equation}
  (\widehat y,z)=E_{\theta_E}(c).
  \label{eq:e-map}
\end{equation}
\textcolor{black}{Forward prediction is also available from the encoder response $\widehat y$ when $c$ is given.}

We use a latent vector $z\in\mathbb R^d$, with fewer
components than the discretized field $c$, to compress information that
complements $y$ for reconstruction. \textcolor{black}{The encoder predicts a mean vector
$\mu\in\mathbb R^d$ and a variance vector $\sigma\in\mathbb R^d$ to construct the latent $z$.
Each component $z_i$ has mean $\mu_i$ and standard deviation
$\sigma_i$. $z$ is constructed as}
\begin{equation}
  z_i=\mu_i+\sigma_i\xi_i,\qquad
  \xi_i\overset{\mathrm{iid}}{\sim}\mathcal N(0,1),\quad i=1,\ldots,d.
  \label{eq:latent-distribution}
\end{equation}
\textcolor{black}{The decoder forms a base prediction and a spatial gate from the response $y$, its validity mask, and coordinates. A separate path injects $z$ at multiple scales and combines it with response features to produce a residual. The two paths give}
\begin{equation}
  \widehat c=D_{\theta_D}(z,y)
    =P_{\mathrm{out}}\!\left(B_{\theta_D}(y)
      +G_{\theta_D}(y)\odot R_{\theta_D}(z,y)\right).
  \label{eq:d-map}
\end{equation}
\textcolor{black}{Here $B_{\theta_D}(y)$ supplies the observation-based prediction, $R_{\theta_D}(z,y)$ is the latent residual, and $G_{\theta_D}(y)$ is the gate, which controls how much the latent residual changes the base prediction. The product $\odot$ is elementwise. $P_{\mathrm{out}}$ maps it to the reconstructed field. We omit the mask and coordinates from the notation for brevity.}
Network and latent-branch details are given in
Appendices~\ref{app:shared-architecture} and~\ref{app:latent-construction}.

Let $\mathcal D_p=\{(c,y^{\mathrm{num}})\}$ be the small set of fields paired with their true PDE responses, and let $\mathcal D_u=\{c\}$ contain additional fields without numerical responses. The decoder condition $\bar y(c)$ follows the task's observation protocol: paired fields can be conditioned on either the numerical response or the predicted response $\widehat y$, while unpaired fields use $\widehat y$. The main autoencoder objective is
\begin{equation}
 \mathcal L_{\mathrm{AE}}
 =\lambda_y\mathcal L_{\mathrm{obs}}
 +\lambda_c\mathcal L_{\mathrm{rec}}
 +\lambda_{\mathrm{phys}}\mathcal L_{\mathrm{phys}}
 +\lambda_{\mathrm{reg}}\mathcal L_{\mathrm{reg}}.
 \label{eq:ae-loss}
\end{equation}
The supervised response loss uses only $\mathcal D_p$; field reconstruction also uses $\mathcal D_u$:
\begin{equation*}
  \begin{aligned}
  \mathcal L_{\mathrm{obs}}
    &=\mathbb E_{(c,y^{\mathrm{num}})\sim\mathcal D_p}
      d_y(\widehat y,y^{\mathrm{num}}),\\
  \mathcal L_{\mathrm{rec}}
    &=\mathbb E_{c\sim\mathcal D_p\cup\mathcal D_u}
      \sum_{q\in\{\mu,z\}}\omega_q(c)\,
      d_c\!\left(D_{\theta_D}(q,\bar y(c)),c\right).
  \end{aligned}
\end{equation*}
Here $d_y$ measures response error, while $d_c$ is a field loss (squared error for continuous fields and binary cross-entropy for piece-wise constant fields). The physics loss is:
\begin{equation*}
  \mathcal L_{\mathrm{phys}}
    =\mathbb E_{(c',y')\sim\mathcal B_{\mathrm{phys}}}
      \left[\frac{1}{|\mathcal W|}\sum_{w\in\mathcal W}
      |r_{\mathrm{PDE}}(c',y';w)|^2\right]
      +\mathcal L_{\mathrm{IC/BC}}.
\end{equation*}
Here $r_{\mathrm{PDE}}(c',y';w)$ is the weak-form PDE residual obtained by testing the governing equation with $w\in\mathcal W$ and integrating by parts in space. The latent regularizer $\mathcal L_{\mathrm{reg}}$ follows the VAE formulation~\citep{kingma2014autoencoding}, pulling the encoder's latent distributions toward a standard Gaussian.

We evaluate weak PDE residuals on the encoder's forward predictions and the decoder's inverse reconstructions. Weak residuals are detailed in Appendix~\ref{app:weak-pde}; autoencoder training (Algorithm~\ref{alg:pae-training}) and loss settings are in Appendix~\ref{app:autoencoder-training} and the task-specific implementation sections.

\subsection{Conditional latent Flow}
\label{sec:flow}
\textcolor{black}{After the autoencoder is trained, its encoder latent output supplies the flow target $z$. When the true observation is available, we use it to condition the Flow. Otherwise, we use the frozen encoder's predicted observation as the condition (Figure~\ref{fig:two-stage-framework}B).}

\textcolor{black}{We use the observation $y$ to guide the evolution of the latent state $z_\tau$ through spatially varying scaling and shifting, inspired by FiLM~\citep{perez2018film}, where $z_\tau=(1-\tau)z_0+\tau z$.}

\textcolor{black}{We train the conditional flow using the standard flow-matching objective~\citep{lipman2023flowmatchinggenerativemodeling,liu2022flowstraightfastlearning}, with straight paths from Gaussian noise to the frozen encoder's latent codes and conditioning on the corresponding observations.}

\textcolor{black}{Flow training is summarized in Algorithm~\ref{alg:flow-training}; architecture, objective, and task-specific conditioning are detailed in Appendices~\ref{app:shared-architecture} and~\ref{app:flow-conditioning}.}

\subsection{Inference}
\label{sec:inference}
\textcolor{black}{For inverse problem, given an observation $y$, we initialize the latent state with Gaussian noise $z_0$ and integrate
$\dot z_\tau=v_{\theta_F}(z_\tau,\tau,y)$
from $\tau=0$ to $\tau=1$ to obtain a latent sample. The decoder uses this latent sample and $y$ to generate the final reconstruction (Figure~\ref{fig:two-stage-framework}C).}

Let $\Phi_{\theta_F}(y;z_0)$ denote the terminal latent sample. To reduce variability due to random sampling, we aggregate reconstructions
from $K$ independent Gaussian sources. Each sample is decoded as:
\begin{equation}
 z^{(k)}=\Phi_{\theta_F}(y;z_0^{(k)}),
 \qquad c^{(k)}=D_{\theta_D}(z^{(k)},y),\quad k=1,\ldots,K.
 \label{eq:sampling}
\end{equation}
The samples from Eq.~\eqref{eq:sampling} give alternative reconstructions for
the observation. A point estimate is obtained using the aggregation rule
specified for each experiment. Algorithm~\ref{alg:pae-flow-inference} summarizes inference.

\paragraph{Optional refinement.}
When observations are noisy, we can refine the response used
to condition reconstruction. Each round updates this response using PDE and
observation errors while keeping any clean-anchor values fixed, then reruns
the frozen Flow and decoder to generate a new latent and reconstruct the
unknown field. Only the response is gradient-optimized; network parameters
remain fixed. The detailed procedure is given in
Appendix~\ref{app:inference-refinement}.

\section{Experiments}
\label{sec:experiments}

\begingroup
We study fine-detail recovery in two types of PDE inverse
problems: recovering an initial field from a later state under Allen--Cahn
dynamics, and recovering a Darcy coefficient from coarse or sparse pressure
measurements. The Darcy experiments cover continuous coefficients and
binary coefficients with sharp interfaces.

\paragraph{Baselines.}
We use parameter-matched No-$Z$ controls that remove the complementary
latent pathway from our AE and adjust decoder widths to match the
parameter budget (Appendix~\ref{app:model-configurations}). PINO~\citep{li2024pino} and
DGenNO~\citep{zang2025dgenno} are evaluated on all three tasks;
DGenNO is included only in the coarse-grid setting without anchors for continuous Darcy.
For Allen--Cahn and continuous Darcy, we also compare with a conditional
EDM~\citep{karras2022elucidating} trained with a PDE-residual loss  (Conditional EDM+PDE) evaluated through the frozen forward branch of our physics-aware
encoder. Additional baselines are
DPS~\citep{chung2023diffusionposterior} for Allen--Cahn and
DiffusionPDE~\citep{huang2024diffusionpde} for binary Darcy.
All methods within each task use the same observations and test fields.
Baseline settings are described in
Appendix~\ref{app:baseline-physics} and
Appendices~\ref{app:ac-implementation}--\ref{app:binary-implementation}.
DGenNO uses a latent size matching OCL-PDE
($512$ for Allen--Cahn and $1024$ for Darcy). We perform direct latent optimization for DGenNO at test time;
Appendix~\ref{app:dgenno-inverse-protocol} compares this procedure with
the inference procedure used in the original paper.
\paragraph{Evaluation metrics.}
For continuous fields, we report relative $L_2$ error,
$\|\widehat c-c\|_2/\|c\|_2$.
For binary Darcy, the primary metric is hard pixel error: the fraction
of grid points assigned the wrong coefficient value.
Additional boundary-recovery metrics are introduced in
Appendix~\ref{app:binary-boundary-metrics}.
We also assess forward consistency by independently solving the PDE
with each reconstructed field and comparing the resulting solution with the reference solution using relative $L_2$ error.
Test time is averaged per held-out field and excludes these independent
numerical checks; for OCL-PDE, it includes generating and decoding all
$K$ samples used for the reconstruction.

\subsection{Backward recovery for the Allen--Cahn equation}
\label{sec:ac-experiment}

\paragraph{Inverse problem and observations.}
On the periodic domain $\Omega=[0,2]^2$, the phase field $\phi(x,t)$
satisfies~\citep{allen1979microscopic}
\begin{equation}
 \begin{aligned}
 \partial_t\phi(x,t)&=\nu\Delta_x\phi(x,t)+\phi(x,t)-\phi(x,t)^3,
 &&x\in\Omega,\ t>0,\\
 \phi(x,0)&=\phi_0(x), &&x\in\Omega.
 \end{aligned}
 \label{eq:ac}
\end{equation}
Our goal is to recover the initial field $\phi_0$ from a later full-field observation.

\begin{figure}[!b]
  \centering
  \includegraphics[width=0.50\textwidth]{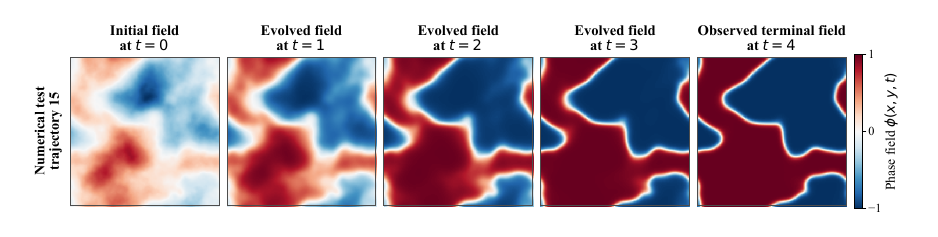}
  \caption{\revisionseven{Allen--Cahn evolution of a trajectory from $t=0$ to $t=4$.}}
  \label{fig:ac-evolution}
\end{figure}

\paragraph{Experimental setup.}
With $\nu=0.0004$, the observation is $\phi(\cdot,4)$; both fields use a $128\times128$ grid. We sample $G\sim\mathcal{GP}(0,(-\Delta+I)^{-5/2})$ spectrally~\citep{li2021fourierneuraloperatorparametric} and normalize each initial field as $\phi_0=G/\|G\|_\infty$. We evolve both true and recovered initial fields to $t=4$ using Fourier spectral discretization and first-order semi-implicit time stepping~\citep{shen2010numerical}, then measure the relative $L_2$ error between the resulting fields.
The latent $z$ has size $8\times8\times8$. At inference, we average $K=64$ decoded samples.
\textcolor{black}{All methods use the same 500 supervised field--response pairs.
Implementation details are given in Appendices~\ref{app:ac-latent}--\ref{app:ac-data-training}.}
Fine details in the initial field are no longer resolved in the terminal observation, making backward recovery difficult as shown in Figure~\ref{fig:ac-evolution}.
\begingroup

\begin{table}[ht]
\centering
\scriptsize
\caption{Allen--Cahn initial-field and terminal-field relative $L_2$ errors on 20 held-out trajectories, with test time per field.}
\label{tab:ac-outline}
\begin{tabular}{@{}lccr@{}}
\toprule
Method & \shortstack{$\phi_0$ relative $L_2$\\mean $\pm$ std.} & \shortstack{Terminal-field relative $L_2$\\at $t=4$; mean $\pm$ std.}  & \shortstack{Test time\\(s/field)} \\
\midrule
\revisionseven{PINO~\citep{li2024pino}} & \revisionseven{$0.18496\pm0.04341$} & \revisionseven{$0.03110\pm0.01316$}  & -- \\
\revisionseven{DGenNO~\citep{zang2025dgenno} ($d=512$)} & \revisionseven{$0.66444\pm0.33967$} & \revisionseven{$0.62601\pm0.20541$}  & $73.51$ \\
\revisionseven{Conditional EDM+PDE~\citep{karras2022elucidating}} & $0.41914\pm0.05692$ & $0.41826\pm0.07390$ & $12.42$ \\
\revisionseven{DPS (L2 guidance)~\citep{chung2023diffusionposterior}} & $0.45889\pm0.14540$ & $0.18284\pm0.10136$ & $27.56$ \\
\revisionseven{No-$Z$ ablation} & $0.09827\pm0.01863$ & $0.01679\pm0.01108$ & -- \\
\revisionseven{OCL-PDE} & $\mathbf{0.05897\pm0.01817}$ & $\mathbf{0.01386\pm0.00547}$ & $0.9150$ \\
\bottomrule
\end{tabular}
\end{table}
\begin{figure}[ht]
  \centering
  \includegraphics[width=0.8\textwidth]{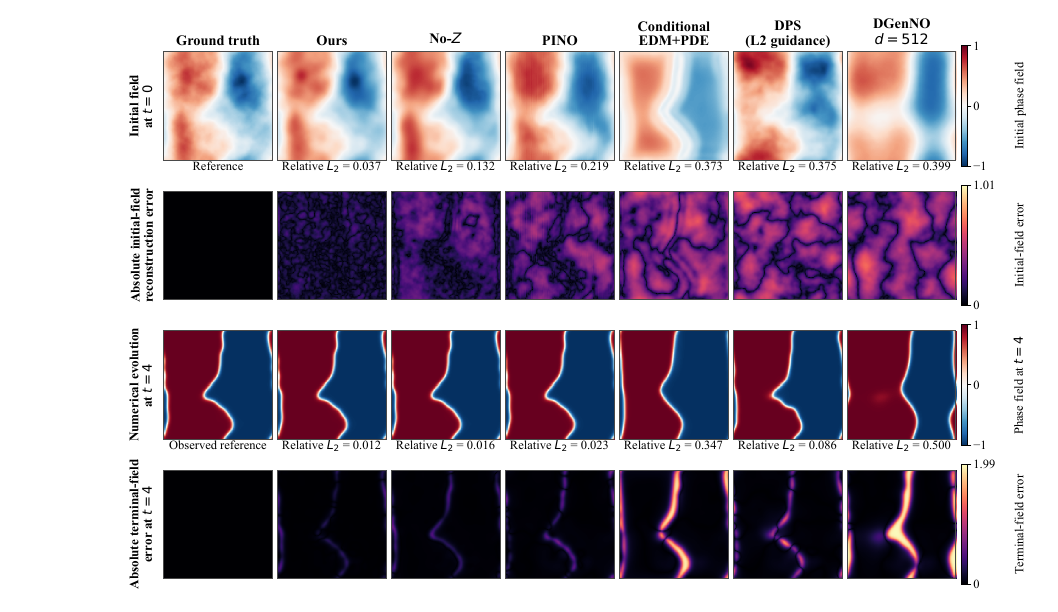}
  \caption{Allen--Cahn initial-field recovery. Rows show reconstructions, absolute initial-field errors, spectral-method evolutions to $t=4$, and absolute terminal-field errors. We report initial-field and terminal-field errors in relative $L_2$.}
  \label{fig:ac-clean-main}
\end{figure}
We recover the Allen--Cahn initial field from its terminal observation $\phi(\cdot,t=4)$ using OCL-PDE and other methods and compare both the reconstructed initial-field and its evolution with their respective ground truths. Initial-field relative $L_2$ error is lowest for OCL-PDE at $0.05897$, compared with $0.18496$ for PINO and $0.09827$ for the parameter-matched No-$Z$ control, which removes the latent pathway (Table~\ref{tab:ac-outline}). The reduction relative to No-$Z$ is about $40\%$. Terminal-field error is also lowest at $0.01386$. OCL-PDE better preserves fine-scale details within each phase, while most other methods mainly recover the overall phase layout
(Figure~\ref{fig:ac-clean-main}). These results support the role of the complementary latent in recovering details unresolved by the terminal observation.

Test time is $0.9150$ s per field for OCL-PDE, compared with $12.42$ s for Conditional EDM+PDE, $27.56$ s for DPS, and $73.51$ s for DGenNO. OCL-PDE samples compact latents; Conditional EDM+PDE samples full fields, DPS and DGenNO repeatedly compute gradients. PINO and No-$Z$ make direct predictions; their timings are omitted from this table. These measurements describe the evaluated inference procedures (Appendix~\ref{app:ac-sampling-cost}). Appendix~\ref{app:ac-forward} separately evaluates the learned forward branch against reference trajectories.
\endgroup

\subsection{Darcy flow with a continuous coefficient and coarse observations}
\label{sec:grf-experiment}

\paragraph{Inverse problem and observations.}
For steady Darcy flow on $\Omega=[0,1]^2$, the pressure
$u(x)$ satisfies
\begin{equation}
 -\nabla_x\!\cdot\bigl(a(x)\nabla_x u(x)\bigr)=f(x),
 \quad x\in\Omega,\qquad u(x)=0,\quad x\in\partial\Omega.
 \label{eq:darcy}
\end{equation}
We recover a spatially varying Darcy coefficient from coarse pressure observations and also test whether exact pressure measurements at selected locations improve recovery when the remaining observations are noisy.

\paragraph{Experimental setup.}
Here $f(x)\equiv1$. We adapt the Gaussian random field generation and sample-wise
normalization used in the Allen--Cahn example to the domain $[0,1]^2$.
On the $128\times128$ grid, the resulting normalized field $g$ defines
the positive Darcy coefficient through $a=\exp(g)$. The continuous-coefficient Darcy experiment uses 1,000 paired fields. 
The latent $z$ has size $4\times16\times16$. We average $K=16$ decoded coefficient samples for the clean-anchor study.
We solve for reference pressures and pressures from recovered coefficients using conservative finite differences with arithmetic face coefficients~\citep{li2021fourierneuraloperatorparametric,huang2024diffusionpde}, the relative $L_2$ error between these pressure  fields is reportedin the $u$ columns of Table~\ref{tab:grf-multifidelity-summary}.

We observe the pressure $u$ on an aligned $32\times32$ grid and recover the coefficient $a$ on a $128\times128$ grid.  After removing
the known zero boundary, the observation contains 900 interior $u(x)$ values.
\textcolor{black}{Implementation details and the training procedure are given in
Appendix~\ref{app:grf-training}.}

We replace a subset of the noisy interior coarse-grid pressure values
with noise-free reference values, called clean anchors.
Writing $\mathcal I_H$ for the anchor set,
\begin{equation}
 y_i^{\mathrm{noisy}}=u_i+\epsilon_i,
 \quad \epsilon_i\sim\mathcal N(0,(\rho s_u)^2),
 \qquad
 y_i^{\mathrm{anchor}}=u_i\ \text{for }i\in\mathcal I_H,
 \qquad r=\frac{|\mathcal I_H|}{900},
 \label{eq:grf-multifidelity}
\end{equation}
where $u_i$ is the noise-free reference pressure at interior node $i$,
$s_u$ is the training-set pressure standard deviation, $\rho$ is
the relative noise level, and $r$ is the anchor fraction among the interior nodes. We use
$\rho\in\{0.01,0.03\}$ and vary $r$ from 0 to 0.5; the known zero boundary remains exact.

\paragraph{Pressure correction with clean anchors.}
We update non-anchor coarse pressure values using PDE and observation gradients,
keeping clean anchors and the zero boundary fixed. The observation loss is
normalized MSE against the fixed noisy measurements, penalizing excessive
deviation from the data. Each update reconstructs the coefficient from the
corrected pressure. We use 20 rounds without a forward solve
(Algorithm~\ref{alg:grf-multifidelity-refinement}).
Appendix~\ref{app:grf-refinement-evaluation} evaluates $0$--$50\%$ clean-anchor
coverage at both noise levels.

\begin{table}[ht]
\centering
\scriptsize
\setlength{\tabcolsep}{2.6pt}
\caption{GRF coefficient reconstruction on 16 test fields with $50\%$
clean-anchor coverage: mean relative $L_2$ errors of the recovered
coefficient ($a$) and corresponding pressure ($u$), before (r0) and
after (r20) pressure correction. Pressure is computed from each
recovered coefficient using conservative finite differences.}
\label{tab:grf-multifidelity-summary}
\begin{tabular*}{\linewidth}{@{\extracolsep{\fill}}l*{8}{r}@{}}
\toprule
Method & \multicolumn{4}{c}{Noise $\rho=1\%$} & \multicolumn{4}{c}{Noise $\rho=3\%$} \\
\cmidrule(lr){2-5}\cmidrule(l){6-9}
 & \shortstack{$a$ r0} & \shortstack{$a$ r20} & \shortstack{$u$ r0} & \shortstack{$u$ r20} & \shortstack{$a$ r0} & \shortstack{$a$ r20} & \shortstack{$u$ r0} & \shortstack{$u$ r20} \\
\midrule
PINO & $0.03836$ & $0.03669$ & $0.00452$ & $0.00450$ & $0.05252$ & $0.03694$ & $0.00580$ & $0.00456$ \\
Conditional EDM+PDE & $0.09808$ & $0.09912$ & $0.02349$ & $0.02398$ & $0.09927$ & $0.09893$ & $0.02372$ & $0.02400$ \\
No-$Z$ ablation & $0.03605$ & $0.03151$ & $0.00345$ & $0.00318$ & $\mathbf{0.05127}$ & $0.04034$ & $0.00531$ & $0.00409$ \\
OCL-PDE & $\mathbf{0.03456}$ & $\mathbf{0.02249}$ & $\mathbf{0.00310}$ & $\mathbf{0.00265}$ & $0.05239$ & $\mathbf{0.02381}$ & $\mathbf{0.00520}$ & $\mathbf{0.00285}$ \\
\bottomrule
\end{tabular*}
\end{table}

\begin{figure}[ht]
  \centering
  \includegraphics[width=0.75\textwidth]{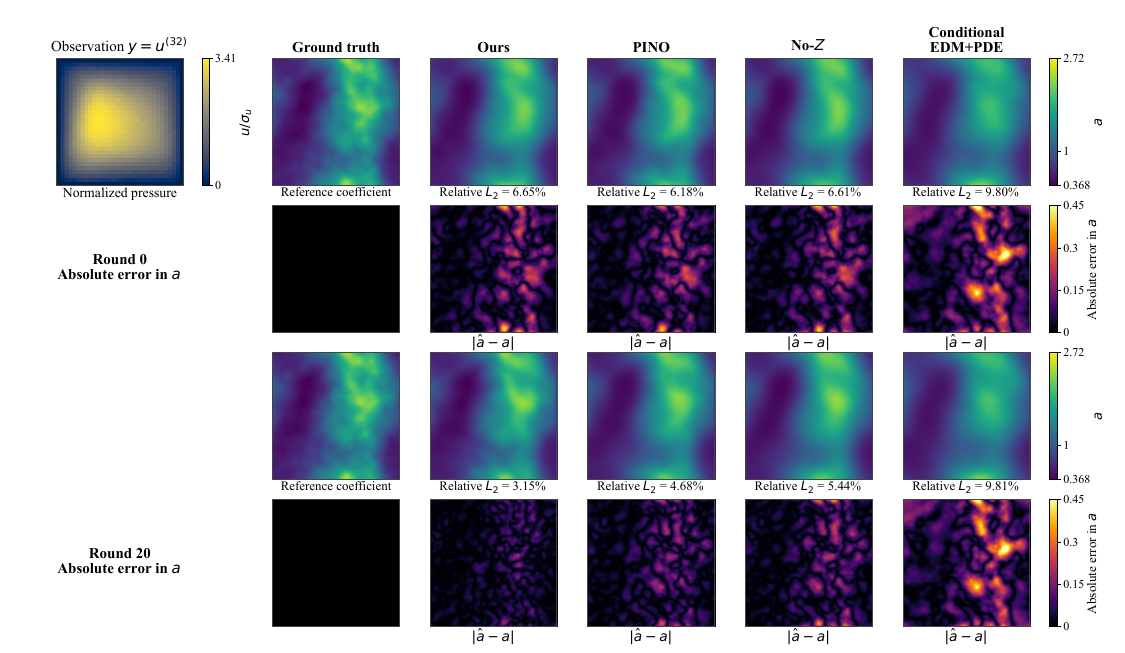}
  \caption{Continuous coefficient recovery before and after pressure correction at $3\%$ noise and $50\%$ clean-anchor coverage. We show the pressure observation, reference coefficient, reconstructions, and absolute coefficient errors, and report coefficient reconstruction errors in relative $L_2$.}
  \label{fig:grf-multifidelity}
\end{figure}
We recover continuous Darcy coefficients from noisy coarse pressures with clean anchors and compare the methods before and after pressure correction. Coefficient error decreases from $0.03456$ to $0.02249$ at $1\%$ noise and from $0.05239$ to $0.02381$ at $3\%$ noise, reductions of $34.9\%$ and $54.6\%$, respectively (Table~\ref{tab:grf-multifidelity-summary}). The largest baseline reduction at $3\%$ noise is $29.7\%$ (PINO). Fine coefficient variations are also recovered more faithfully after correction in the illustrated field (Figure~\ref{fig:grf-multifidelity}). These results show that OCL-PDE makes more effective use of clean anchors to improve reconstruction accuracy, while the evaluated baselines benefit less from these high-quality observations.

Appendix~\ref{app:grf-noise-robustness} tests noise robustness without clean anchors.
Pressure error is lowest for OCL-PDE at every tested noise level.
Coefficient relative $L_2$ error remains controlled, rising from $0.01807$
without noise to $0.05740$ at $3\%$ noise, and is comparable to or below the evaluated baselines.
\subsection{Darcy flow with a piecewise constant coefficient}
\label{sec:binary-experiment}

\paragraph{Inverse problem and observations.}
The pressure follows Eq.~\eqref{eq:darcy}, with constant forcing
$f(x)\equiv1$ and a two-valued coefficient $a(x)\in\{3,12\}$.
Our goal is to recover the piecewise constant coefficient from sparse pressure measurements.

\paragraph{Experimental setup.}
We sample $G\sim\mathcal{GP}(0,(-\Delta+9I)^{-2})$ in a cosine basis without the constant mode and set $a(x)=3+9\mathbf{1}_{\{G(x)\geq0\}}$, following Darcy benchmarks~\citep{li2021fourierneuraloperatorparametric,zang2025dgenno}. We recover $a$ on a
$128\times128$ grid from random interior pressure observations at
$2\%$, $5\%$, and $10\%$ observation rates.
Training uses 1,000 paired fields,
supplemented by dynamically generated coefficients during latent training.
We encode the coefficient geometry as a truncated signed-distance
field (TSDF)~\citep{curless1996volumetric} to construct a latent of size
$4\times16\times16$ (Appendix~\ref{app:binary-latent}).
We average $K=16$ phase-probability samples and threshold at $0.5$.
The same conservative finite-difference solver computes pressures from the
recovered hard coefficients, table~\ref{tab:binary-random-main}'s $u$ columns
report their relative $L_2$ errors against reference pressures.
\textcolor{black}{Implementation and training details are given in
Appendix~\ref{app:binary-data-training}.}

\begin{table}[ht]
\centering
\scriptsize
\setlength{\tabcolsep}{0.7pt}
\caption{Binary coeficient reconstruction on 64 test fields:
hard pixel error of the recovered coefficient, relative $L_2$ error
of the corresponding pressure, and inference time (seconds per field).
Pressure is computed from each recovered hard coefficient using
conservative finite differences.}
\label{tab:binary-random-main}
\begin{tabular*}{\linewidth}{@{\extracolsep{\fill}}l*{9}{r}@{}}
\toprule
Method & \multicolumn{3}{c}{2\% (318 points)} & \multicolumn{3}{c}{5\% (794 points)} & \multicolumn{3}{c}{10\% (1588 points)} \\
\cmidrule(lr){2-4}\cmidrule(lr){5-7}\cmidrule(l){8-10}
 & \textcolor{black}{Hard error $\downarrow$} & \textcolor{black}{$u$ rel. $L_2$ $\downarrow$} & \textcolor{black}{Time} & \textcolor{black}{Hard error $\downarrow$} & \textcolor{black}{$u$ rel. $L_2$ $\downarrow$} & \textcolor{black}{Time} & \textcolor{black}{Hard error $\downarrow$} & \textcolor{black}{$u$ rel. $L_2$ $\downarrow$} & \textcolor{black}{Time} \\
\midrule
PINO & $0.06800$ & $0.08264$ & - & $0.05070$ & $0.06317$ & - & $0.04572$ & $0.05456$ & - \\
DGenNO ($d=1024$) & $0.13957$ & $0.10923$ & $99.02$ & $0.14084$ & $0.11891$ & $96.24$ & $0.14459$ & $0.12381$ & $84.42$ \\
\shortstack[l]{DiffusionPDE\\\citep{huang2024diffusionpde}} & $0.15472$ & $0.06302$ & $164.8$ & $0.12350$ & $0.05192$ & $164.5$ & $0.10744$ & $0.04561$ & $164.6$ \\
No-$Z$ & $0.04659$ & $0.02709$ & - & $0.03454$ & $0.01950$ & - & $0.03143$ & $0.01753$ & - \\
OCL-PDE & $\mathbf{0.03524}$ & $\mathbf{0.01919}$ & $0.3417$ & $\mathbf{0.02357}$ & $\mathbf{0.01199}$ & $0.3437$ & $\mathbf{0.02105}$ & $\mathbf{0.01086}$ & $0.3440$ \\
\bottomrule
\end{tabular*}
\end{table}

\begin{figure}[ht]
  \centering
  \includegraphics[width=\textwidth]{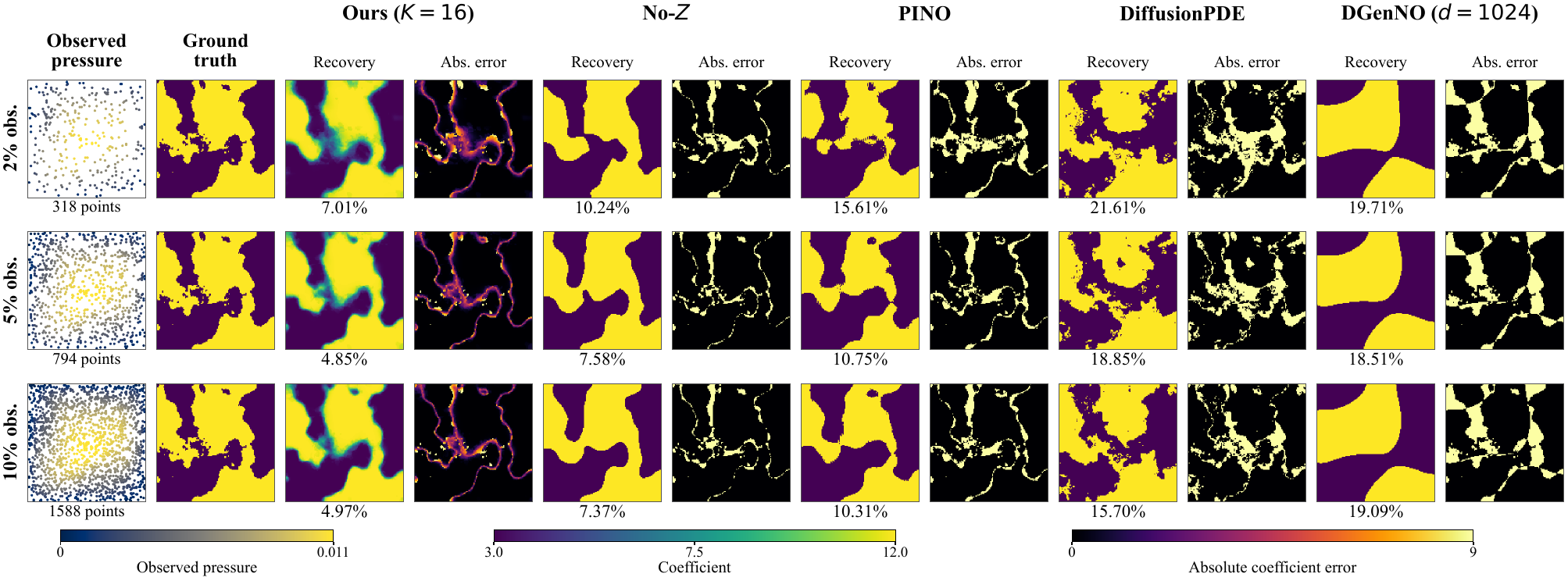}
  \caption{Binary coefficient recovery on the same held-out field at $2\%$, $5\%$, and $10\%$ observation rates (top to bottom). Percentages below reconstructions are hard pixel errors.}
  \label{fig:binary-random-main}
\end{figure}

We compare binary coefficient recovery and forward consistency at three observation densities. Hard pixel error and pressure relative $L_2$ error are lowest for OCL-PDE at every density (Table~\ref{tab:binary-random-main}). At $5\%$ observations, hard pixel error is $2.36\%$, compared with $5.07\%$ for PINO, a $53.5\%$ relative reduction. Error is also $31.8\%$ lower than the No-$Z$ ablation's $3.45\%$. Recovered region shapes and small holes match the reference more closely (Figure~\ref{fig:binary-random-main}), supporting the latent's role in recovering sharp interfaces from sparse observations.

Test time is approximately $0.34$ s per field for OCL-PDE, $84$--$99$ s for DGenNO,
and $165$ s for DiffusionPDE with 2,000 guided diffusion steps.
Boundary F1 is highest and ASSD and HD95 are lowest at all densities,
supporting improved interface detail recovery (Appendix~\ref{app:binary-results}).
Learned forward prediction is also most accurate (Appendix~\ref{app:binary-forward}).
Without retraining or refinement, hard pixel error remains stable at 80 and 50 dB
observation noise, and OCL-PDE outperforms No-$Z$ and PINO at $2\%$, $5\%$,
and $10\%$ observation density (Appendix~\ref{app:robustness-diagnostics}).
{\color{black}{
\subsection{Ablation study: what does the complementary latent contribute?}
\label{sec:ablations}

\paragraph{Does the latent improve reconstruction?}
OCL-PDE improves reconstruction over parameter-matched No-$Z$ controls,
which use only the observation pathway, across all three problems
(Tables~\ref{tab:ac-outline},
\ref{tab:grf-multifidelity-summary}, and~\ref{tab:binary-random-main}).
These gains demonstrate the benefit of the latent beyond the observation pathway alone.

\paragraph{Are $y$ and $z$ complementary?}
For Allen--Cahn, we exchange either the observation or the latent between
two samples. The coarse arrangement and outlines of the two phase regions follow the observation: swapping $y$ changes which phase
occupies a given region. The spatial patterns within each phase follow
the latent, changing when $z$ is replaced while $y$ remains fixed
(Figure~\ref{fig:ac-cross-pair}; further interventions in
Appendix~\ref{app:latent-evidence}).
These swaps support the complementary roles of $y$ and $z$: $y$ primarily determines the phase arrangement, while $z$ supplies fine-scale details within the phases.

\paragraph{What does the learned latent recover?}
We fix the observation and set $z=0$ in the frozen decoder to obtain
a base reconstruction. The latent correction follows spatial patterns
missing from this base, partially recovering their locations and signs
(Figure~\ref{fig:latent-grf-single}; Appendix
Figures~\ref{fig:latent-correction-grf}--\ref{fig:latent-correction-binary}).
Positive $Q$ confirms that these corrections reduce the base's pooled
squared error. Lower $E_{\mathrm{hi}}$ further shows improved agreement
with the ground truth at high frequencies
(Table~\ref{tab:latent-correction-metrics}, Appendix~\ref{app:latent-corrections}).
Thus, the latent restores missing spatial structure, including a portion
of its fine-scale variation.}}

\endgroup 
\begingroup
\setlength{\intextsep}{4pt plus 1pt minus 1pt}
\begin{figure}[!htb]
  \setlength{\abovecaptionskip}{3pt}
  \centering
  \includegraphics[width=0.7\textwidth]{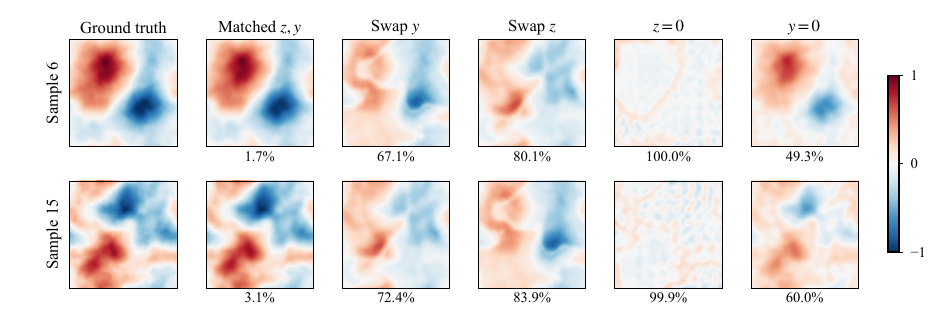}
  \caption{Allen--Cahn reconstructions for one pair of samples, swapping $y$ or $z$ between rows or setting either input to zero. Errors are relative $L_2$ against each row's ground truth.}
  \label{fig:ac-cross-pair}
\end{figure}
\begin{figure}[!htb]
  \setlength{\abovecaptionskip}{3pt}
  \centering
  \includegraphics[width=0.7\textwidth]{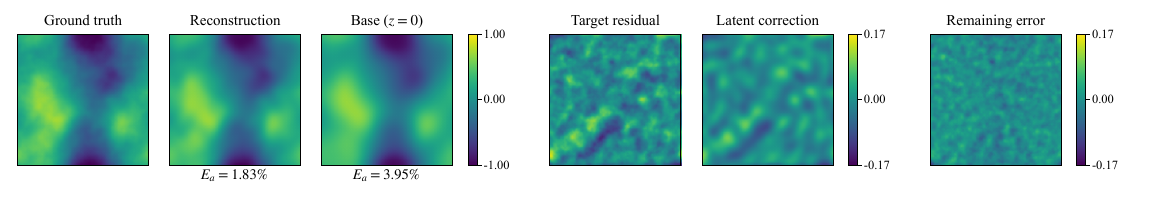}
  \caption{GRF Darcy latent correction in $\log a$ (Sample 0). The last three columns show what the base ($z=0$) misses, what the latent adds, and what remains unrecovered, respectively. Coefficient errors are relative $L_2$.}
  \label{fig:latent-grf-single}
\end{figure}

\section{Conclusion and Future Work}

We presented OCL-PDE, which combines a physics-aware autoencoder with
conditional Flow Matching to learn and generate observation-complementary
latents for PDE inverse problems. Experiments on backward Allen--Cahn initial state
recovery and continuous and binary Darcy coefficient inversion demonstrate improved
reconstruction accuracy and detail recovery over the evaluated baselines.
Ablations confirm that the latent supplies information beyond the
observation, with task-dependent complementary roles.

Beyond initial-state recovery and coefficient inversion, future work will explore extending OCL-PDE to other PDE inverse problems, such as recovering heat-source distributions from temperature measurements and identifying unknown boundary conditions from interior measurements.
\endgroup

\section*{AI Use Statement}
We used large language models to assist with literature search,
translation, manuscript revision, method implementation
and debugging, experiment execution, and interpretation of experimental results.
The authors reviewed the AI-assisted work and verified the references,
implementation, and reported results. The authors take full
responsibility for the final manuscript, implementation, and findings.

\section*{Ethics Statement}
This work studies PDE inverse problems using synthetically generated
physical fields and numerical simulations. It does not involve human
participants, personal data, or animal experiments. We do not identify
specific ethical concerns arising from the experiments reported here.

\section*{Reproducibility Statement}
The appendix provides details of the network architectures, data generation,
training settings, and evaluation protocols.

\section*{Acknowledgements}
The work of D.Y., C.M. and Y.X. was supported in part by the National Key Research and Development Program of China (No. 2025YFA1016800) and the Project of Hetao Shenzhen-HKUST Innovation Cooperation Zone HZQB-KCZYB-2020083.

\bibliographystyle{iclr2027_conference}
\bibliography{pae_flow_references}

\appendix
\section{Architecture, training, and inference details}
\label{app:architecture-implementation}

Throughout the appendix, \emph{GRF Darcy} denotes the
continuous-coefficient Darcy-flow experiment, in which the log-permeability
field is generated from a Gaussian random field (GRF).

\subsection{Shared network architecture}
\label{app:shared-architecture}

\paragraph{Physics-aware autoencoder.}
All three tasks use the same four-part design rather than a single CNN or a
plain U-Net.  A shallow convolutional stem is shared by an FNO-based
deterministic response branch~\citep{li2021fourierneuraloperatorparametric}
and a convolutional latent branch.
The response branch predicts the observable or forward field, whereas the
latent branch produces the compact complementary target $z$.  A U-Net-style
conditional decoder, inspired by the contracting/expanding organization and
skip connections of U-Net~\citep{Ronneberger_2015}, reconstructs the unknown
field from the predicted or measured response, its validity mask, coordinates,
and $z$.  \textcolor{black}{The observation path produces both the base prediction
$B_{\theta_D}(y)$ and a spatial gate $G_{\theta_D}(y)$. In parallel, the latent
path projects $z$ at multiple decoder scales and combines it with observation
features to form the correction $R_{\theta_D}(z,y)$. As in
Eq.~\eqref{eq:d-map}, their outputs combine as}
\begin{equation}
  {\color{black}\widehat c=D_{\theta_D}(z,y)
    =P_{\mathrm{out}}\!\left(B_{\theta_D}(y)
      +G_{\theta_D}(y)\odot R_{\theta_D}(z,y)\right).}
  \label{eq:appendix-decoder-map}
\end{equation}
\textcolor{black}{The base supplies the observation-based estimate, and the gate
controls the latent correction at each location. The sum is formed before the
task-specific output mapping $P_{\mathrm{out}}$, which is the identity for
Allen--Cahn; the Darcy mappings are specified in their implementation sections.
The latent correction path is an addition to the original U-Net.}

\paragraph{Conditional latent Flow.}
\textcolor{black}{The velocity network $v_{\theta_F}$ takes the current latent state $z_\tau$, Flow time $\tau$, and an observation condition. Its spatial channels are $\eta_y=[\widetilde y,M_y,A_y]$, where $\widetilde y$ is the normalized observation (zero-filled where needed), $M_y$ marks valid measurements, and $A_y$ contains optional task-specific channels. A convolutional extractor $C_{\theta_F}$ maps $\eta_y$ to features $h_y$ at the latent resolution. An input convolution maps $z_\tau$ to the first hidden state $h_0$. In residual convolutional block $l$~\citep{he2016residual}, a projection $P_l$ of $h_y$ produces a spatially varying scale and shift:}
\begin{equation}
  {\color{black}\begin{aligned}
    h_y&=C_{\theta_F}(\eta_y),\qquad
    (\gamma_y^{(l)},\beta_y^{(l)})=P_l(h_y),\\
    \operatorname{Mod}_l(h_l;y)
      &=(1+\gamma_y^{(l)})\odot\operatorname{GN}(h_l)
        +\beta_y^{(l)}.
  \end{aligned}}
  \label{eq:appendix-flow-modulation}
\end{equation}
\textcolor{black}{Here $h_l$ is the block's hidden state, $\operatorname{GN}$ is group normalization, and $\operatorname{Mod}_l$ is spatial FiLM-style conditioning~\citep{perez2018film}. Flow time $\tau$ is embedded sinusoidally~\citep{vaswani2017attention,ho2020ddpm} and adds channel-wise modulation; Allen--Cahn also embeds its known observation time $t_{\mathrm{cond}}=4$. One four-head self-attention block~\citep{vaswani2017attention} separates two stacks of residual blocks. After the final block, group normalization, SiLU activation, and an output convolution produce $v_{\theta_F}(z_\tau,\tau,y)$ with the same shape as $z_\tau$.}

\textcolor{black}{Allen--Cahn and Binary Darcy use multi-scale condition extractors inspired by the lateral projections of feature pyramid networks~\citep{Lin_2017_CVPR}; GRF Darcy uses a single downsampling stage from its native observation grid. We call the multi-scale extractors \emph{FPN-inspired} because they aggregate projected features at the latent resolution without the top-down pathway of the original FPN. Task-specific condition channels and Flow widths are given in Table~\ref{tab:network-io-conditioning} and the implementation sections below.}

\subsection{Latent construction}
\label{app:latent-construction}
All tasks use direct full-field encoding. The latent branch
receives a representation $h=T(c)$ of the complete unknown field: the initial
phase field for Allen--Cahn, log-permeability for GRF Darcy, and a scaled
signed-distance field for Binary Darcy. For both Darcy tasks, we center
the latent mean at the zero input,
$\mu=E^z_{\theta_E}(h)-E^z_{\theta_E}(0)$, where $E^z_{\theta_E}$ denotes the
uncentered latent-mean branch. This fixes the latent reference at $h=0$.
The encoder supplies $(\mu,\ell)$, and sampling follows
Eq.~\eqref{eq:latent-distribution}. Task-specific transforms and latent sizes
are given in Appendices~\ref{app:ac-latent}--\ref{app:binary-latent}.
The latent encodes the complete transformed field $T(c)$, not its difference
from a zero-latent reconstruction. In every task, Flow is conditioned on
the observation and the task-specific inputs listed in
Table~\ref{tab:network-io-conditioning}; $D(0,y)$ is not a Flow condition.
The decoder's latent correction pathway and the zero-latent diagnostic
remain distinct from this full-field encoding.

\subsection{Weak-form physics losses}
\label{app:weak-pde}
We construct weak-form PDE losses by the method of weighted
residuals~\citep{finlayson1972weighted}: multiply the governing equation
by a test function $w$ and integrate by parts in space. For example,
applying this procedure to the Allen--Cahn and Darcy equations used
in our experiments gives
\begin{align}
  r_{\mathrm{AC}}(\phi;w)
  &=\int_\Omega\!\left[\partial_t\phi\,w
    +\nu\nabla\phi\cdot\nabla w+(\phi^3-\phi)w\right]\,d\boldsymbol x,
    \label{eq:ac-weak-residual}\\
  r_{\mathrm{Darcy}}(a,u;w)
  &=\int_\Omega\!\left[a\nabla u\cdot\nabla w-fw\right]\,d\boldsymbol x.
    \label{eq:darcy-weak-residual}
\end{align}
Here $\phi$ and $u$ denote the forward solution $y$ for Allen--Cahn and
Darcy, respectively; $a=c$ is the Darcy coefficient, and $f$ is its forcing.
The compactly supported test functions below eliminate the spatial
boundary term. These residuals require only first spatial derivatives of
the response and no spatial derivatives of $a$.

For predicted responses in all three tasks, we use translated Wendland
$C^2$ compactly supported radial basis functions
(CSRBFs)~\citep{wendland1995piecewise,bao2025pfwnn,zang2025dgenno},
\begin{equation}
  w_{\boldsymbol\xi}(\boldsymbol x)
  =\psi\!\left(\frac{\|\boldsymbol x-\boldsymbol\xi\|_2}{R}\right),
  \qquad \psi(r)=(1-r)_+^4(4r+1),
  \label{eq:csrbf-test}
\end{equation}
where $(r)_+=\max(r,0)$, $\boldsymbol\xi$ is a test center, and $R=4h$
is the support radius, with $h$ denoting the spatial sampling interval.
Each $w_{\boldsymbol\xi}\in H_0^1(\Omega)$ vanishes outside the ball
$B(\boldsymbol\xi,R)$, so its residual can be evaluated using only a local
neighborhood. The radius and discrete normalization here
follow our training implementation.
On the $128\times128$ training grids, we set $h=2/127$ for Allen--Cahn and
$h=1/128$ for both Darcy tasks.  With support radius $R=4h$, each local weak
residual is evaluated over a $9\times9$ neighborhood.  We repeat this
calculation at all grid points whose neighborhoods lie entirely within the
domain.  Spatial response derivatives are computed by second-order
finite differences on the predicted fields. For Allen--Cahn, the temporal
derivative is computed by automatic differentiation in the shared physical-time
direction. Test-function derivatives are evaluated analytically.  The CSRBF loss is
the mean squared discrete residual over training samples and valid centers.
Darcy uses quadrature-weighted sums with area factor $h^2$, whereas
Allen--Cahn uses the unscaled local sums; these conventions determine the
normalization of their physics-loss weights.

In both Darcy tasks, the CSRBF terms constrain predicted pressure fields
with either supplied or reconstructed coefficients.  During AE training,
we also constrain reconstructed coefficients using fixed numerical pressure
fields.  This auxiliary weak PDE loss uses the same CSRBF test
functions in Eq.~\eqref{eq:csrbf-test}. Pressure gradients are computed by
centered finite differences. Each squared local residual is normalized by
the squared local integral of $fw_{\boldsymbol\xi}$, with a lower bound of
$10^{-20}$ on the denominator, and the loss averages over valid test
centers and training samples. These are autoencoder-stage losses; the
subsequent Flow stage uses only the velocity objective in Eq.~\eqref{eq:appendix-straight-fm}.

\subsection{\textcolor{black}{Autoencoder training}}
\label{app:autoencoder-training}

\paragraph{Optimization stages.}
\textcolor{black}{Autoencoder optimization follows task-specific schedules for
the response branch, observation-based decoder path, and latent correction
path (Appendices~\ref{app:ac-implementation}--\ref{app:binary-implementation}).}

\paragraph{Auxiliary variance head.}
The latent encoder produces the mean $\mu_z$.
An auxiliary $1\times1$ convolutional head predicts
$\ell_z=H_{\mathrm{var}}(\mu_z)=\log\sigma_z^2$ elementwise.
Together they define the Gaussian distribution in
Eq.~\eqref{eq:latent-distribution}. Both belong to $\theta_E$ in the main
text; the ``Log-var'' column of Table~\ref{tab:appendix-parameter-counts}
counts the head separately for parameter accounting.

\paragraph{\textcolor{black}{Latent sampling and KL regularization.}}
For all three PDE examples, initial Gaussian latent learning
uses both the mean $\mu$ and sampled latents
\begin{equation}
  z^\star=\mu+\alpha\sigma\odot\xi,\qquad
  \xi\sim\mathcal N(0,I).
  \label{eq:latent-sample}
\end{equation}
Here $\odot$ denotes elementwise multiplication and $\alpha$
controls the sampling spread; $\alpha=0$ gives the mean. During initial
latent learning, $\alpha$ is annealed up to $0.5$. The KL regularizer is
averaged over latent components and training samples:
\begin{equation}
  \mathcal L_{\mathrm{reg}}=\mathcal L_{\mathrm{KL}}
  =\mathbb E\!\left[\frac{1}{2d}\sum_{i=1}^{d}
  \left(\mu_i^2+\sigma_i^2-\log\sigma_i^2-1\right)\right].
  \label{eq:latent-regularization}
\end{equation}
Its coefficient reaches $10^{-4}$ at full weight.

\paragraph{\textcolor{black}{Autoencoder training algorithm.}}
\textcolor{black}{Algorithm~\ref{alg:pae-training} summarizes autoencoder
training. Task-specific observation protocols, loss weights, and schedules
are given in Appendices~\ref{app:ac-implementation}--\ref{app:binary-implementation}.}

\begin{algorithm}[H]
  \caption{\textcolor{black}{Autoencoder training}}
  \label{alg:pae-training}
  \small
  {\color{black}
  \begin{algorithmic}[1]
    \Require Training fields $c$ and paired numerical responses when available
    \Ensure Encoder $E_{\theta_E}$ and decoder $D_{\theta_D}$
    \For{each autoencoder update}
      \State Sample fields and available paired responses; obtain $\widehat y$ from $E_{\theta_E}$ with input $c$.
      \State Form $y$ from predicted or numerical responses using the observation protocol.
      \State Encode the complete field $c$ to obtain $(\mu,\ell)$ using Appendix~\ref{app:latent-construction}; set $\sigma=\exp(\ell/2)$.
      \State Form $z^\star=\mu+\alpha\sigma\odot\xi$ with $\xi\sim\mathcal N(0,I)$.
      \State Reconstruct $c$ from both $(\mu,y)$ and $(z^\star,y)$ using $D_{\theta_D}$.
      \State Update the scheduled subsets of $(\theta_E,\theta_D)$ using Eq.~\eqref{eq:ae-loss}.
      \Statex \hspace{\algorithmicindent}Evaluate $\mathcal L_{\mathrm{obs}}$ only where paired responses are available.
    \EndFor
  \end{algorithmic}}
\end{algorithm}

\subsection{\textcolor{black}{Conditional Flow training and conditioning}}
\label{app:flow-conditioning}
\textcolor{black}{The autoencoder is frozen throughout Flow training.}

\paragraph{\textcolor{black}{Flow targets and normalization.}}
\textcolor{black}{Flow targets use $\alpha=0.5$ for Allen--Cahn and Binary Darcy,
and $\alpha=0$ for both GRF observation settings, so GRF targets are encoder
means. This does not remove the Gaussian source used for Flow inference.
Allen--Cahn uses source standard deviation $\sigma_0=0.04$ and no latent
standardization. Darcy uses $\sigma_0=1$ and standardizes target latents with
per-channel means and standard deviations estimated from training data; this
transform is inverted before decoding. These fixed transforms are included in
$\Phi_{\theta_F}$ in the main text. Allen--Cahn additionally pairs source and
target samples by minibatch optimal transport before velocity matching.}

\paragraph{\textcolor{black}{Flow objective.}}
\textcolor{black}{We train the Flow with straight conditional Flow Matching
(equivalently, the one-step rectified-flow objective)~\citep{lipman2023flowmatchinggenerativemodeling,liu2022flowstraightfastlearning}:}
\begin{equation}
  {\color{black}z_\tau=(1-\tau)z_0+\tau z_1,
  \qquad
  v^\star=z_1-z_0,
  \qquad
  \mathcal L_{\mathrm{FM}}
  =\mathbb E\!\left[\lVert v_{\theta_F}(z_\tau,\tau,y)-v^\star\rVert_2^2\right].}
  \label{eq:appendix-straight-fm}
\end{equation}
\textcolor{black}{ ``One-step'' here denotes one
rectification stage in the learning objective, not one numerical integration
step: we do not perform iterative reflow, and all formal Flow evaluations
below use 50 Euler steps.}

\paragraph{\textcolor{black}{Flow training algorithm.}}
\textcolor{black}{Algorithm~\ref{alg:flow-training} summarizes Flow training.
Task-specific observation protocols and schedules are given in
Appendices~\ref{app:ac-implementation}--\ref{app:binary-implementation}.}

\begin{algorithm}[H]
  \caption{\textcolor{black}{Conditional Flow training}}
  \label{alg:flow-training}
  \small
  {\color{black}
  \begin{algorithmic}[1]
    \Require Frozen trained encoder $E_{\theta_E}$, training fields $c$, and paired numerical responses when available
    \Ensure Velocity network $v_{\theta_F}$
    \For{each Flow update}
      \State Form $(\mu,\ell,y)$ using the frozen autoencoder and observation protocol.
      \State Form $z^\star$ using the stated sampling scale and convert it to the Flow target $z_1$.
      \State Draw $z_0\sim\mathcal N(0,\sigma_0^2I)$ and $\tau\sim\mathcal U[0,1]$.
      \State Apply the stated source--target pairing and set $z_\tau=(1-\tau)z_0+\tau z_1$.
      \State Update $\theta_F$ using Eq.~\eqref{eq:appendix-straight-fm}.
    \EndFor
  \end{algorithmic}}
\end{algorithm}

\paragraph{\textcolor{black}{Conditioning channels.}}
The condition extractor $C$ is part of the Flow velocity network $v_{\theta_F}$.
It converts the observation channels into spatial features on the latent grid.
The velocity network uses these features, the current latent $z_\tau$, and
Flow time $\tau$ to predict a velocity with the same shape as $z_\tau$.
The task-specific observation channels are collected in
Table~\ref{tab:network-io-conditioning} and explained in the implementation
sections below. We write $\eta(y)$ for these channels to distinguish them
from the unknown physical field $c$.

\begin{table}[H]
  \centering
  \scriptsize
  \setlength{\tabcolsep}{3.5pt}
  \begin{tabularx}{\textwidth}{@{}lYll@{}}
    \toprule
    Task & Spatial condition supplied to $C$ & Latent/velocity shape
      & Scalar inputs \\
    \midrule
Allen--Cahn
      & $\eta_{\mathrm{AC}}=[\phi_4,\mathbf 1]$ on $128^2$
      & $8\times8\times8$ & $\tau$ and $t_{\mathrm{cond}}=4$ \\
GRF Darcy, no anchors
      & $\eta_{\mathrm{low}}=[u/s_u,M]$ on $32^2$
      & $4\times16\times16$ & $\tau$ only \\
GRF Darcy, clean anchors
      & $\eta_{\mathrm{MF}}=[u/s_u,M,M_H]$ on $32^2$
      & $4\times16\times16$ & $\tau$ only \\
Binary Darcy & $\eta_{\mathrm{bin}}=[y_q/s_u,M_q]$ on $128^2$ & $4\times16\times16$ & $\tau$ only \\
\bottomrule
  \end{tabularx}
  \caption{Spatial inputs to the Flow condition extractor $C$.
  Here $\phi$ is the Allen--Cahn solution and
  $\phi_4=\phi(\boldsymbol x,4)$ is its observed terminal field ($y$);
  $\mathbf1$ is the all-one validity mask.
  For GRF Darcy, $u$ denotes the native $32^2$ pressure observation; for Binary Darcy, $y_q$ is the raw zero-filled $128^2$ pressure tensor. The quantity $s_u$ is the training-set pressure standard deviation. In the clean-anchor row, selected noisy values in $u$ have been replaced with clean anchors. The masks $M$ and $M_q$ mark observed pressure nodes, and $M_H$ marks clean anchors. In GRF Darcy, Flow receives native-grid observations.
  Flow time $\tau$ and physical time $t_{\mathrm{cond}}$ are separate
  scalar inputs; the noise level $\rho$ is not supplied to any Flow.}
  \label{tab:network-io-conditioning}
\end{table}

\subsection{Inference and optional refinement}
\label{app:inference-refinement}
Forward prediction uses the response output $\widehat y$ of
$E_{\theta_E}(c)$. Algorithm~\ref{alg:pae-flow-inference} describes inverse
sampling: the frozen Flow generates latents conditioned on the observation,
and the frozen decoder reconstructs the unknown field. Optional refinement
updates the response used as the condition under PDE and observation
constraints, while keeping trusted measurements and network weights fixed.
Reconstruction is then repeated with the same Gaussian source samples.
The continuous-Darcy refinement procedure is specified in
Appendix~\ref{app:grf-multifidelity-refinement}.

\begin{algorithm}[H]
  \caption{Inverse sampling and optional refinement}
  \label{alg:pae-flow-inference}
  \small
  \begin{algorithmic}[1]
    \Require Observation $y$, frozen $D_{\theta_D}$ and $v_{\theta_F}$, and sample count $K$
    \Ensure Reconstructed fields and an optional point estimate
    \For{$k=1,\ldots,K$}
      \State Draw $z_0^{(k)}\sim\mathcal N(0,\sigma_0^2I)$ independently.
      \State Integrate $\dot z_\tau=v_{\theta_F}(z_\tau,\tau,y)$ from $\tau=0$ to $1$.
      \State Undo any latent normalization to obtain $z^{(k)}=\Phi_{\theta_F}(y;z_0^{(k)})$.
      \State Decode $c^{(k)}=D_{\theta_D}(z^{(k)},y)$.
    \EndFor
    \State If needed, form a point estimate using the specified aggregation rule.
    \State If refinement is used, update the condition and repeat reconstruction with fixed sources,
    \Statex \hspace{\algorithmicindent}following Algorithm~\ref{alg:grf-multifidelity-refinement}.
    \State \Return the reconstructed fields and the optional point estimate.
  \end{algorithmic}
\end{algorithm}

\section{Experiment Settings}
\label{app:evaluation-protocols}

\subsection{Model configurations and parameter counts}
\label{app:model-configurations}

\paragraph{Parameter counts.}
Frozen components remain loaded during inference.  Accordingly, the
totals in Table~\ref{tab:appendix-parameter-counts} count all parameters loaded
by the audited inference graph, not parameters trained simultaneously in one
optimization stage.

For Allen--Cahn, PINO's forward and inverse networks total 52.74M
parameters ($+0.02\%$ relative to OCL-PDE's 52.73M). DPS uses
41,699,617 prior parameters and 11,167,993 forward-operator parameters,
totaling 52,867,610 (52.87M), a $+0.26\%$ difference from OCL-PDE.

For GRF Darcy, PINO and the No-$Z$ clean-anchor model have 40.58M and
40.70M parameters, respectively ($-0.37\%$ and $-0.09\%$ relative to
OCL-PDE's 40.73M). These differences use the recorded components for each
procedure; no uniform matching tolerance is claimed for all methods.

For Binary Darcy, the audited PINO, No-$Z$, DGenNO ($d=1024$), and
DiffusionPDE models contain 40.58M, 40.91M, 40.75M, and 41.92M parameters,
respectively, within $-0.58\%$ to $+2.71\%$ of OCL-PDE's 40.82M.

\begin{table}[ht]
  \centering
  \resizebox{\textwidth}{!}{%
  \begin{tabular}{@{}lcccccc@{}}
    \toprule
    Task & Field grid & Latent $C\times H\times W$ & AE & Log-var & Flow & Total \\
    \midrule
Allen--Cahn & $128\times128$ & $8\times8\times8$
      & 30,650,596 & 72 & 22,080,456 & 52,731,124 \\
GRF Darcy & $128\times128$ & $4\times16\times16$
      & 30,313,272 & 20 & 10,417,444 & 40,730,716 \\
Binary Darcy & $128\times128$ & $4\times16\times16$ & 30,313,272 & 20 & 10,503,668 & 40,816,960 \\
\bottomrule
  \end{tabular}%
  }
  \caption{Audited parameter counts for the models used in formal inference.
  The GRF Flow with clean anchors has one additional clean-anchor-mask condition
  channel.}
  \label{tab:appendix-parameter-counts}
\end{table}

\subsection{Baseline architectures and PDE losses}
\label{app:baseline-physics}

\paragraph{PINO forward and inverse networks.}
Across Allen--Cahn, continuous Darcy, and binary Darcy, each
PINO baseline consists of two separate FNO networks: a forward network
$F_{\mathrm{PINO}}:c\mapsto y$ and an inverse network
$G_{\mathrm{PINO}}:y\mapsto c$. For Darcy these maps are $a\mapsto u$ and
$u_{\mathrm{obs}}\mapsto a$; for Allen--Cahn they map the initial field to a
later field and the observed later field back to the initial field.
Both networks belong to the PINO baseline and are trained independently
of the OCL-PDE autoencoder. All network weights are frozen at evaluation.

For continuous Darcy, the inverse inputs are normalized
observed pressure, an observation mask, forcing, and the two spatial
coordinates. With clean anchors, the mask is replaced by the
validity-plus-anchor status channel. The forward FNO is frozen during
inverse training. Its pressure prediction initializes the first
clean-anchor refinement step; later steps retain the accepted pressure iterate.

\paragraph{Baseline PDE losses.}
PINO uses the same CSRBF weak-residual formulation as OCL-PDE
for Allen--Cahn and both Darcy tasks, including GRF settings with and
without clean anchors. In the shared grid-based implementation,
second-order finite differences evaluate spatial field derivatives inside
the weak residual; Allen--Cahn time derivatives use automatic differentiation
in the physical-time direction. Conditional EDM+PDE uses a CSRBF loss on the predicted
response: pressure for Darcy and the learned forward response for
Allen--Cahn. AC DPS retains its unconditional prior and uses the retrained
forward operator for observation and CSRBF PDE guidance. Binary DiffusionPDE
uses a grid finite-difference PDE residual during inference. DGenNO uses
CSRBF weak residuals with response derivatives evaluated at independent
query points. Task-specific training and inference settings are given in
Appendices~\ref{app:ac-implementation}--\ref{app:binary-implementation}.

\subsection{DGenNO inverse inference protocols}
\label{app:dgenno-inverse-protocol}
\paragraph{Direct-latent inference used in our comparisons.}
We freeze the trained DGenNO generators and optimize a separate latent
variable $\beta$ for each test field. We use eight initializations:
the training-code mean, observation-matched training codes when a code
bank is available, and uniformly sampled codes to fill the remaining
starts. Matching uses only the available response observations and their mask.
For each initialization, we run 500 Adam steps on latent coordinates
scaled by their training-code standard deviations, with projection to
the latent bounds, followed by at most 50 L-BFGS iterations.

The objective combines observation mismatch and a weak PDE-residual loss
using 32 residual centers; Allen--Cahn additionally includes initial-condition
consistency between the generated initial field and response branch.
We retain the lowest-objective candidate encountered throughout all
iterations and restarts, without using the test field's ground-truth
reconstruction error for selection. The resulting single $\beta^*$ is
decoded into the reconstructed coefficient or initial field.

This direct-latent multistart procedure differs from the original
Beta-MLP parameterization, which also optimizes the latent representation
at test time. In Binary Darcy, we threshold the selected generator's
probability at $p\geq0.5$ to obtain the hard coefficient field used for
visualization and hard pixel error.

\paragraph{Original Beta-MLP inference procedure.}
We separately evaluate the original method's Beta-MLP parameterization on
our checkpoints and test fixtures. The network has three ReLU hidden layers
of widths 64, 64, and 128, followed by a $d$-dimensional Tanh output.
One fit uses 1,000 Adam steps with weight decay $10^{-4}$, learning rate $10^{-2}$ divided by three
every 250 steps, 100 weak-residual centers, and 25 integration points.
We set $\lambda_{\mathrm{PDE}}=1$ and $\lambda_{\mathrm{data}}=50$ for
Binary Darcy or 25 for continuous fields.
The implementation returns a single optimized $\beta^*$.
This is a paper-aligned adaptation, not an exact reproduction: checkpoints,
coefficient values, observation geometries, and test fixtures differ from
the original experiments. Allen--Cahn is a task adaptation outside the
original paper. All times below exclude independent numerical verification;
entries are mean $\pm$ sample standard deviation.
The task-specific appendix tables report this paper-aligned Beta-MLP
inference using the same frozen checkpoints as the direct-latent evaluations.

\subsection{Additional Metrics for Binary Darcy}
\label{app:binary-boundary-metrics}

Results using these definitions are reported in
Table~\ref{tab:binary-random-main} for hard pixel error and solver-consistency
error, and in Appendix~\ref{app:binary-results},
Table~\ref{tab:binary-random-boundary}, for Boundary F1, ASSD, and HD95 at
$2\%$, $5\%$, and $10\%$ observation rates.

Tolerance-based boundary F-measures compare contour precision and recall while
allowing small localization
errors~\citep{martin2004learning,cheng2021boundary}. Nearest-surface
distance summaries are likewise standard for segmentation evaluation, and a
Hausdorff quantile reduces sensitivity to isolated outliers
~\citep{taha2015metrics}.  We specify our exact pixel-grid conventions below
because symmetrization, tolerance, and empty-set handling vary across
implementations.

The Binary-Darcy metrics are computed per held-out field on the
$H\times W=128\times128$ grid, then averaged arithmetically across fields.
Let $S$ denote the true binary phase mask: at pixel $p$, $S(p)=1$ if
$a(p)=12$ and $S(p)=0$ if $a(p)=3$.  The predicted mask $\widehat S$ is
defined analogously from the reconstructed hard coefficient
$\widehat a(p)\in\{3,12\}$.  Hard pixel error (HardErr) is the fraction of all grid pixels whose predicted coefficient value differs from the true value.  The independent solver-consistency error is
\begin{equation}
  \operatorname{SolverErr}
  =\frac{\|\mathcal F(\widehat a)-\mathcal F(a)\|_2}
             {\|\mathcal F(a)\|_2},
  \label{eq:binary-region-solver-metrics}
\end{equation}
where $\mathcal F$ is an independent numerical Darcy solver using the
benchmark forcing and boundary conditions.  SolverErr measures the relative
pressure error induced by the reconstructed hard coefficient.

For any binary mask $S$, its boundary-pixel set is
\begin{equation}
  B(S)=\left\{p:
  \max_{q\in\mathcal N_3(p)}S(q)
  \ne
  \min_{q\in\mathcal N_3(p)}S(q)
  \right\},
  \label{eq:binary-boundary-set}
\end{equation}
where $\mathcal N_3(p)$ is the centered $3\times3$ pixel neighborhood.
For Binary Darcy, $p$ is a boundary pixel if this neighborhood contains
both phases, i.e., pixels with coefficient values $a=3$ and $a=12$;
the same rule applies to the predicted coefficient $\widehat a$.
Thus $B(S)$ is a narrow band around the two-phase interface, whereas $S$
labels the entire regions.

Let $\mathcal D_\delta(B)$ be the set of grid pixels whose absolute row and
column offsets from at least one boundary pixel in $B$ are both at most
$\delta$.  For the reported tolerance $\delta=1$, $\mathcal D_1(B)$ is the
union of the centered $3\times3$ neighborhoods of all pixels in $B$,
including $B$ itself.  A predicted boundary pixel is counted as correctly
localized if at least one true boundary pixel lies within one row and one
column of it; equivalently, it lies in $\mathcal D_1(B(S))$.  This rule
tolerates a small location error without requiring exact overlap or
one-to-one pairing of boundary pixels.  Boundary F1 is then
\begin{equation}
  P_1=\frac{|B(\widehat S)\cap\mathcal D_1(B(S))|}
            {\max(|B(\widehat S)|,1)},
  \qquad
  R_1=\frac{|B(S)\cap\mathcal D_1(B(\widehat S))|}
            {\max(|B(S)|,1)},
  \qquad
  \operatorname{BF1}_1=\frac{2P_1R_1}{\max(P_1+R_1,10^{-8})}.
  \label{eq:binary-boundary-f1}
\end{equation}
Here $P_1$ is the fraction of predicted boundary pixels close to a true
boundary, and $R_1$ is the fraction of true boundary pixels covered by a nearby
predicted boundary.  ``Close'' allows a one-pixel offset horizontally,
vertically, or diagonally.  Extra predicted edges reduce precision, while
missed true edges reduce recall; their harmonic mean rewards agreement in
both directions.  Boundary F1 ranges from zero to one, with higher values
indicating better boundary agreement, not a higher fraction of correctly
classified region pixels.  The maxima in the denominators avoid division by zero.

For the distance metrics, let
$d(p,B)=\min_{q\in B}\lVert p-q\rVert_2$ in pixel units and form the
bidirectional nearest-boundary distance vector
\begin{equation}
  \Delta(\widehat S,S)=
  \bigl(d(p,B(S))\bigr)_{p\in B(\widehat S)}
  \mathbin{\Vert}
  \bigl(d(q,B(\widehat S))\bigr)_{q\in B(S)}.
  \label{eq:binary-boundary-distances}
\end{equation}
The first vector contains one distance per predicted boundary pixel to its
closest true boundary pixel; the second contains one distance per true
boundary pixel to its closest predicted boundary pixel.  Here $\Vert$
denotes concatenation: the two vectors are placed end to end, not maximized
or added elementwise, and repeated distances are retained.  For example,
$(0,1,1)\mathbin{\Vert}(0,2)=(0,1,1,0,2)$.
A horizontal or vertical one-pixel offset has distance $1$, while a diagonal
one-pixel offset has distance $\sqrt{2}$.  We summarize the combined vector as
\begin{equation}
  \operatorname{ASSD}=\operatorname{mean}(\Delta),
  \qquad
  \operatorname{HD95}=Q_{0.95}(\Delta).
  \label{eq:binary-assd-hd95}
\end{equation}
ASSD measures the average boundary displacement, giving equal weight to
every entry in the combined vector.  HD95 summarizes larger local boundary
mismatches: approximately $95\%$ of the distances fall at or below this value.
Using this percentile rather than the single largest distance reduces the
influence of a few isolated extreme errors.  Both metrics are measured in
pixels, with lower values indicating better alignment.  As a hypothetical
single-field example, ASSD $=4$ px means that the bidirectional nearest-boundary
distances average $4$ pixels, not that every boundary pixel is displaced by
$4$ pixels.  HD95 $=4$ px instead means that approximately $95\%$ of these
distances are at most $4$ pixels; the remaining tail may be larger.
These metrics are calculated
for each field before averaging across fields; the reported HD95 is therefore
the mean of field-wise percentiles, not a percentile pooled over all fields.
If either boundary set is empty,
both distances are assigned the finite worst-case image diagonal
$\sqrt{H^2+W^2}$ pixels rather than infinity or a missing value.

\section{Allen--Cahn implementation and additional results}
\label{app:ac-implementation}
\subsection{Latent construction}
\label{app:ac-latent}
For Allen--Cahn, the unknown field is the initial state $c=\phi_0$.
The latent branch encodes this initial field directly:
\begin{equation}
  (\widehat\phi_t,\mu_z,\ell_z)=E_{\theta_E}(\phi_0),\qquad \mu_z\in\mathbb R^{8\times8\times8}.
  \label{eq:ac-latent-design}
\end{equation}
The latent computation uses the encoder's shared convolutional stem. The latent contains
eight channels on an $8\times8$ grid. It is learned jointly with a decoder
that also receives the terminal observation, encouraging it to retain
information useful for initial-field reconstruction.

\subsection{Data, networks, and training}
\label{app:ac-data-training}

\paragraph{Initial-field generation.}
On the periodic domain $[0,2]^2$, we generate
$G\sim\mathcal{GP}(0,(-\Delta+I)^{-5/2})$ by spectral sampling with the
zero Fourier mode removed, and normalize each sample as
$\phi_0=G/\|G\|_\infty$. This follows the spectral Gaussian-process
construction used in neural-operator benchmarks~\citep{li2021fourierneuraloperatorparametric},
with the covariance and normalization specified here.

\paragraph{Network architecture.}
Both the initial field and terminal observation are represented on a
$128\times128$ grid, and the latent tensor has shape $8\times8\times8$.  The
forward branch contains four Fourier blocks of width 104 with $8\times8$
retained modes.  Each block adds a spectral convolution to a pointwise
$1\times1$ convolution and applies GroupNorm~\citep{wu2018groupnorm} and
SiLU~\citep{elfwing2018silu}; the output head uses a
$\tanh$ activation.  The latent CNN contains three shallow
convolutional/residual feature stages followed by four stride-2 downsampling
blocks.

The decoder has four downsampling stages, a bottleneck, and four upsampling
stages with skip concatenations.  Its channel schedule is
$64/96/128/192/256$.  The parallel latent correction path injects projected
$z$ features at full, $1/2$, $1/4$, $1/8$, and bottleneck resolutions before
the gated merge with the base reconstruction.  Decoder blocks use GroupNorm,
SiLU, reflection padding, and bilinear upsampling.  The Flow condition module
is a four-scale FPN-inspired extractor with lateral $1\times1$ projections;
the projected scales are pooled and fused at the $8\times8$ latent
resolution, without a top-down pathway.  The Flow backbone has width 256,
five spatially modulated residual blocks before and after one four-head
self-attention block.

\paragraph{Encoder, decoder, and Flow inputs.}
Let $B$ denote the batch size. The AE receives the initial field
$\phi_0\in\mathbb R^{B\times1\times128\times128}$ and the query map
$[x_1,x_2,t]\in\mathbb R^{B\times3\times128\times128}$.  Its response branch
returns $\widehat\phi_t\in\mathbb R^{B\times1\times128\times128}$, and its
latent branch returns a mean of shape $B\times8\times8\times8$.  The decoder
receives $z$, the terminal field (predicted $\phi_t$ during AE training or
observed $\phi_4$ at inference), an all-one validity mask, and
$[x_1,x_2,t=4]$, and reconstructs
$\phi_0\in\mathbb R^{B\times1\times128\times128}$.  The Flow maps
$z_\tau\in\mathbb R^{B\times8\times8\times8}$ and the two-channel field
$[\phi_4,\mathbf 1]\in\mathbb R^{B\times2\times128\times128}$, together
with $\tau\in\mathbb R^{B\times1}$ and
$t_{\mathrm{cond}}=4\in\mathbb R^{B\times1}$, to a velocity of the same
shape as $z_\tau$.  Parent-Flow training uses the frozen AE prediction of
$\phi_4$ from each fresh GRF draw; calibration and formal testing use the raw
numerical $\phi_4$.  The mask remains all ones.  No training-standard-
deviation normalization is applied because the field is already
approximately in $[-1,1]$.  Throughout, the formal condition is
$t=4$ and $\phi_4$.

\paragraph{Training and evaluation settings.}
The AE is optimized for 50,000 updates with AdamW~\citep{loshchilov2019adamw}
at peak learning rate
$3\times10^{-4}$.  The parent Flow is optimized for 30,000 updates at
$10^{-4}$.  Formal numerical-$\phi_4$ calibration then updates only the
195,264 parameters of the Flow condition encoder for 200 updates at
$10^{-5}$; the remaining Flow parameters stay frozen.  These stages use
weight decay $10^{-4}$ and cosine decay~\citep{loshchilov2017sgdr} to $10^{-6}$.  Training uses seed 0.

The AE uses \textcolor{black}{500 numerical field--response pairs across observation times} and a fresh online
GRF draw at every update.  With the AE frozen, the parent Flow constructs its
training pairs dynamically from fresh GRFs.  Calibration \textcolor{black}{reuses 100 of these numerical}
$\phi_4$ conditions with an 80/20 split generated using seed 20260715.  The
formal test uses the 20 fixed held-out trajectories, $K=64$ Flow samples, 50
Euler steps, and stochastic-sampling seed 9091.

\subsection{Iterative inference and timing}
\label{app:ac-sampling-cost}
Conditional EDM+PDE trains for 20 epochs using only the
denoising loss, then linearly increases $\lambda_{\mathrm{PDE}}$ from zero
to 100 over the next 20 epochs and holds it fixed thereafter. The objective
is $10\mathcal L_{\mathrm{EDM}}+64\lambda_{\mathrm{PDE}}\mathcal L_{\mathrm{PDE}}$.
The PDE residual uses the frozen forward operator, spatial field differences,
and a physical-time Jacobian--vector product. TF32 is disabled.

For the AC comparison, Conditional EDM+PDE uses the
final raw weights after 110 training epochs, $K=64$ samples, and
25 Karras--Heun steps. The sampler evaluates the denoiser twice per step
except at the final step, giving 49 evaluations per sample; it runs without
test-time gradients. 

DPS uses one chain with 1,000 reverse-diffusion steps per field.
Its L2 observation loss is backpropagated through the denoiser and learned
forward operator. The weak PDE term has weight 1 and is enabled during the
last $20\%$ of the chain, when the observation term is weighted by 0.1.
The sampler applies a guidance step size of 0.01 to the combined gradient.
DPS uses the retained unconditional prior and the retrained
forward operator. Its mean sampling time is $27.56$ s per field, excluding
the independent numerical rollout used to compute FDM error. 

DGenNO uses eight starts, 500 Adam steps per start, and up to
50 L-BFGS iterations. 

OCL-PDE uses $K=64$ latent samples and 50 Euler steps,
followed by decoding and averaging. The reported times are recorded
inference costs for these procedures, rather than a controlled comparison
with identical sample counts and timer boundaries.

\subsection{DGenNO with the original inference procedure}
\label{app:ac-dgenno-replay}
We adapt DGenNO's original inverse inference procedure to recover
Allen--Cahn initial fields. The original paper~\citep{zang2025dgenno}
reports no Allen--Cahn experiment, so our task-specific training provides
a reference comparison (Table~\ref{tab:dgenno-replay-ac}).
\begin{table}[htbp]
\centering\scriptsize
\setlength{\tabcolsep}{3pt}
\caption{Paper-aligned Allen--Cahn task adaptation ($N=20$). Errors are relative $L_2$. The discrepancy between learned and independently solved terminal responses indicates a forward-surrogate limitation.}
\label{tab:dgenno-replay-ac}
\begin{tabular}{@{}lccccc@{}}
\toprule
Method & $d$ & Initial $\phi_0$ & Learned response & Solver response & Time (s) \\
\midrule
DGenNO & 512 & $1.57088\pm0.51192$ & $0.36875\pm0.20238$ & $0.67583\pm0.33643$ & $16.67510\pm3.37500$ \\
\bottomrule
\end{tabular}
\end{table}

\subsection{forward prediction}
\label{app:ac-forward}
Our learned response operator $E_y(\phi_0,t)$ takes time
as an input, so one checkpoint represents the continuous-time forward map
$\phi_0\mapsto\phi_t$. We evaluate the retrained response operator at
$t\in\{1,2,3,4\}$ using mean per-trajectory relative $L_2$ error over
the same 20 held-out trajectories.
EDM and DPS define no learned forward operator.
OCL-PDE has the lowest forward relative $L_2$ error at all four evaluated times.
\begin{center}
\vspace{-0.5\baselineskip}
\begingroup
\revisionsevencolor
\scriptsize
\setlength{\tabcolsep}{3.5pt}
\begin{tabular}{@{}lcccc@{}}
\toprule
Method & \multicolumn{4}{c}{Relative $L_2$ error $\downarrow$} \\
\cmidrule(l){2-5}
& $t=1$ & $t=2$ & $t=3$ & $t=4$ \\
\midrule
PINO & $0.011347$ & $0.013178$ & $0.017165$ & $0.02221$ \\
DGenNO ($d=512$) & $0.36088$ & $0.41163$ & $0.45939$ & $0.49121$ \\
No-$Z$ ablation & $0.010124$ & $0.010537$ & $0.011881$ & $0.014101$ \\
OCL-PDE & $0.009851$ & $0.009926$ & $0.011308$ & $0.013504$ \\
\bottomrule
\end{tabular}
\endgroup
\vspace{-0.5\baselineskip}
\end{center}

\section{GRF-coefficient Darcy implementation and additional results}
\label{app:grf-implementation}
\subsection{Latent construction}
\label{app:grf-latent}
For continuous Darcy, the unknown field is the permeability $c=a$, and
the networks represent it as $g=\log a$.
The latent branch directly encodes $g$ and centers its mean
using the zero field:
\begin{equation}
  \mu_z=E^z_{\theta_E}(g)-E^z_{\theta_E}(0),\qquad
  \mu_z\in\mathbb R^{4\times16\times16}.
  \label{eq:module-latent-interface}
\end{equation}
The decoder retains the observation-conditioned base $g_0(y)$.
The same latent construction is used with and without clean anchors.

\subsection{Data, networks, and training}
\label{app:grf-training}

\paragraph{Network architecture.}
The permeability and pressure fields use $128\times128$ grids and the latent
tensor has shape $4\times16\times16$.  The deterministic Darcy branch has five
Fourier blocks of width 80 with $10\times10$ modes.  Each block combines a
spectral and pointwise convolution, uses GELU~\citep{hendrycks2016gelu}
between blocks, and uses SiLU in
the projection head.  A parameter-free analytic Dirichlet envelope multiplies
the output to impose the zero pressure boundary exactly~\citep{lagaris1998artificial}.  The latent CNN has
three convolutional/residual stages and three stride-2 downsampling blocks.
The four-level U-Net-style decoder uses channels
$64/96/128/192/256$ and a parallel five-scale latent correction path, with
GroupNorm, SiLU, reflection padding, and bilinear upsampling.

For both coarse-grid $M=32$ observation settings, the Flow condition
extractor uses one stride-2 stage and a $1\times1$ projection to produce
features at the $16\times16$ latent resolution. The Flow backbone has width 192, four
spatially modulated residual blocks, one four-head self-attention block, and
four further residual blocks.

\paragraph{Encoder and decoder inputs.}
The hidden representation is $g=\log a$.  The response operator receives
$g\in\mathbb R^{B\times1\times128\times128}$, coordinates
$[x_1,x_2]\in\mathbb R^{B\times2\times128\times128}$, and forcing
$f\in\mathbb R^{B\times1\times128\times128}$, and outputs the full pressure
field in $\mathbb R^{B\times1\times128\times128}$.  The latent mean has shape
$B\times4\times16\times16$.  Given $z$, condition pressure, validity mask,
coordinates, and forcing, the decoder returns raw $g$ logits on $128^2$; the
physical postprocessing is $g=\tanh(D(\cdot))$ and $a=\exp(g)$.

\paragraph{Flow conditioning.}
The Flow receives native $32\times32$ pressure observations,
their validity mask $M$, and, when applicable, a clean-anchor mask $M_H$.
Pressure is divided by the training-set standard deviation $s_u$.
Without clean anchors,
\begin{equation}
 \eta_{\mathrm{low}}=[u_{32}/s_u,M],
 \label{eq:grf-low-condition}
\end{equation}
while replacing selected noisy values with exact measurements gives
\begin{equation}
 \eta_{\mathrm{MF}}=[u_{\mathrm{mix}}/s_u,M,M_H].
 \label{eq:grf-mf-condition}
\end{equation}
Both conditions stay on the native $32^2$ grid and contain
no noise-level token. The decoder retains its $128^2$ inputs, obtained by
endpoint-aligned nodal bilinear interpolation of pressure with a dense
interior validity mask.
The Flow is trained with anchor fractions from $0$--$50\%$ and encoder-mean
targets; inference starts from Gaussian latent sources.

\paragraph{Training and evaluation settings.}
AE training is sequential: we first train the forward branch for 100 epochs,
then the observation-conditioned base decoder with $z=0$ for 30 epochs,
and finally introduce $z$ to reconstruct the full coefficient field
for 80 epochs.  Each epoch has 100 optimizer updates,
giving 21,000 updates, with respective peak learning rates
$10^{-3}$, $2\times10^{-4}$, and $3\times10^{-4}$.  The Flow runs for
$80\times100=8{,}000$ updates at $2\times10^{-4}$.  All stages use AdamW,
weight decay $10^{-4}$, and cosine decay to $10^{-6}$.  The training set
contains 1,000 numerical pairs and the held-out set contains 64 fields.  Each
AE update combines a paired batch with an online GRF batch.  Flow conditions
are drawn approximately equally from paired numerical pressures and dynamic
GRFs passed through the frozen forward model.

OCL-PDE, PINO, No-$Z$, and Conditional EDM+PDE each use the same model
for low-fidelity and clean-anchor observations. Model parameters remain
frozen during evaluation and pressure refinement. DGenNO is evaluated only
in the setting without clean anchors.

\textcolor{black}{After these AE stages, we fine-tune the latent encoder and
decoder's latent correction path for 40 epochs, with 100 updates per epoch.
The forward branch, observation-based base decoder, and variance head remain
frozen. Each update reconstructs the same coefficient field from two
independently sampled pressure observations at a shared noise level.
Alongside these reconstruction losses, the reference penalty
$5\times10^{-4}\operatorname{mean}[((\mu-\mu^{\mathrm{ref}})/s)^2]$
keeps the adapted latent mean close to the frozen source encoder's mean;
$s$ is the source mean's per-channel RMS over batch and spatial dimensions,
floored at $0.05$. Its weight ramps up over the first five epochs.
Because both observations encode the same complete field $g$, the
paired-latent consistency term is identically zero. Validation selects
epoch 39. The selected AE is frozen while training the single
mixed-condition Flow, and the resulting AE--Flow pair is used for both
low-fidelity and clean-anchor evaluations.}

Conditional EDM+PDE is trained for 110 epochs, sampling anchor
coverage from $0$--$50\%$ and noise on the remaining observations from $0$--$3\%$.

OCL-PDE uses 50 Euler steps in both evaluations, with
$(N,K)=(64,64)$ for low-fidelity observations and $(16,16)$ for clean-anchor
refinement. Field, noise, Flow, and anchor-order seeds are 49091, 59091,
69091, and 820812, respectively. The latter evaluation runs 20 refinement
rounds; its full protocol is given in Appendix~\ref{app:grf-refinement-evaluation}.

\subsection{GRF pressure refinement with clean anchors}
\label{app:grf-multifidelity-refinement}
Refinement alternates between correcting the native pressure $u$ using PDE
and observation errors, and reconstructing $z$ and $a$ from the corrected
condition using the frozen Flow and decoder.  Only $u$ is updated by
gradients; model parameters, $z$, and $a$ are not gradient-optimized.
Algorithm~\ref{alg:grf-multifidelity-refinement} summarizes the procedure for
one field.  It uses the encoder $E_{\theta_E}$, decoder $D_{\theta_D}$, and Flow endpoint
map $\Phi_{\theta_F}$ from \textcolor{black}{Section~\ref{sec:inference}}; $k$ indexes refinement
rounds and $j$ indexes the $K$ fixed Gaussian sources.  The condition $y_k$
uses normalized native pressure and masks for the Flow, and lifted
pressure with the decoder validity mask, as described with
Eq.~\eqref{eq:grf-mf-condition}.

\begin{algorithm}[H]
  \caption{GRF pressure refinement with clean anchors for OCL-PDE}
  \label{alg:grf-multifidelity-refinement}
  \small
  \begin{algorithmic}[1]
    \Require Noisy pressure $u_{\mathrm{noisy}}$, clean-anchor values $u_{\mathrm{clean}}$ and mask $M_H$,
      frozen $(E_{\theta_E},D_{\theta_D},v_{\theta_F})$, sample count $K$, and round count $R$
    \Ensure Refined pressure and coefficient $(u,a)$
    \State Set $u_0=(1-M_H)u_{\mathrm{noisy}}+M_Hu_{\mathrm{clean}}$ and
      $u_0\vert_{\partial\Omega}=0$.
    \State Draw fixed sources $z_0^{(j)}\sim\mathcal N(0,I)$, $j=1,\ldots,K$.
    \State Form $y_0$ from $u_0$ and compute $z_{1,0}^{(j)}=\Phi_{\theta_F}(y_0;z_0^{(j)})$.
    \State Reconstruct the initial coefficient
      $a_0=K^{-1}\sum_{j=1}^K D_{\theta_D}(z_{1,0}^{(j)},y_0)$.
    \State Compute $u_{\mathrm{enc},0}=\operatorname{Down}_{32}\!\left(E_y(\log a_0,\boldsymbol x,f)\right)$;
      impose zero-boundary and clean-anchor values.
    \For{$k=0,\ldots,R-1$}
      \State Start from $u_{\mathrm{enc},0}$ if $k=0$, otherwise from $u_k$;
        enforce the pressure constraints.
      \State Hold $a_k$ fixed; use the losses in Eq.~\eqref{eq:grf-refine-losses}
        to compute $d_k$ by Eq.~\eqref{eq:grf-refine-direction}.
      \State Find an accepted $u_{k+1}$ by projected backtracking along $d_k$.
      \If{no step is accepted}
        \State \Return $(u_k,a_k)$.
      \EndIf
      \State Form the updated condition $y_{k+1}$ from $u_{k+1}$.
      \State Update $z_{1,k+1}^{(j)}=\Phi_{\theta_F}(y_{k+1};z_0^{(j)})$, $j=1,\ldots,K$.
      \State Update $a_{k+1}=K^{-1}\sum_{j=1}^K
        D_{\theta_D}(z_{1,k+1}^{(j)},y_{k+1})$.
    \EndFor
    \State \Return $(u_R,a_R)$.
  \end{algorithmic}
\end{algorithm}

\paragraph{Initialization and reconstruction.}
The initial coefficient $a_0$ is reconstructed from the mixed observation
$u_0$ by the Flow and decoder before the encoder is used.  Here $E_y$ is the
response branch of $E_{\theta_E}$, and $\operatorname{Down}_{32}$ restricts its
pressure prediction to the native $32\times32$ observation grid.
We use $K=16$ standard Gaussian sources and 50 Euler steps for each Flow
integration, reusing the same sources across all rounds and anchor fractions. Subsequent rounds
start from the previously accepted pressure $u_k$; later encoder predictions
are diagnostics only and never replace this iterate.

\paragraph{Pressure losses and update direction.}
We optimize the current pressure values $u_i$ on the native $32\times32$
observation grid, while the original noisy measurements
$u_{\mathrm{noisy},i}$ remain fixed. Let $F$ denote the interior non-anchor
nodes where updates are allowed; clean anchors and the zero boundary are
held fixed.
With $a_k$ fixed and interpolated bilinearly to this grid,
$R_{\mathrm{FV}}(u;a_k,f)$ is the discrete strong-form Darcy residual
$-\nabla\!\cdot(a_k\nabla u)-f$. We evaluate it with a conservative
five-point finite-volume stencil: each face coefficient is the arithmetic
mean of its two neighboring nodal coefficients, and pressure differences
across neighboring nodes determine the face fluxes~\citep{eymard2000finite}.
The PDE loss averages squared residuals after excluding the two outermost
grid layers. The observation loss penalizes deviations of the current
pressure from the original noisy measurements at the free nodes:
\begin{equation}
  \mathcal L_P(u;a_k)=
    \operatorname{mean}\!\left(R_{\mathrm{FV}}(u;a_k,f)^2\right),
  \qquad
  \mathcal L_D(u)=\operatorname{mean}_{i\in F}
    \left(\frac{u_i-u_{\mathrm{noisy},i}}{s_u}\right)^2.
  \label{eq:grf-refine-losses}
\end{equation}
Thus, $\mathcal L_D$ is a normalized MSE between the current coarse-grid
pressure and the fixed noisy data. It acts as a data-fidelity penalty,
discouraging the PDE-driven update from moving too far from the available
measurements. Clean-anchor values are imposed as hard constraints rather
than included in this loss. The first update starts from the constrained encoder prediction;
later updates start from the previously accepted pressure, without
replacing it by a new forward prediction.
The residual is evaluated using physical pressure $u$, while the optimization
variable is normalized pressure $u/s_u$. Writing $\nabla_F$ for this gradient
restricted to free nodes, we normalize the two components separately by their
root mean square (RMS), combine them with weights $0.95$ and $0.05$, and
normalize the result once more:
\begin{equation}
  \widetilde d_k=-\left(
    0.95\frac{\nabla_F\mathcal L_P}{
      \operatorname{RMS}_F(\nabla_F\mathcal L_P)}
    +0.05\frac{\nabla_F\mathcal L_D}{
      \operatorname{RMS}_F(\nabla_F\mathcal L_D)}\right),
  \qquad
  d_k=\frac{\widetilde d_k}{\operatorname{RMS}_F(\widetilde d_k)},
  \label{eq:grf-refine-direction}
\end{equation}
with safe denominators for zero components.  No gradient passes through the
coefficient reconstruction or any network during this pressure step.

\paragraph{Constraints, acceptance, and stopping.}
Before computing the direction and after proposing each step, we enforce
zero boundary values, exact clean anchors, and the trust constraint
$\operatorname{RMS}_F((u-u_{\mathrm{noisy}})/s_u)\le\rho$.
The initial step size in normalized pressure is $\eta_0=0.05\rho$;
in physical units, a candidate adds $s_u\eta d_k$ to the constrained starting
pressure.  Backtracking tests this step and at most 20 successive halvings,
accepting only a strict decrease in
\begin{equation}
  \mathcal J_k(u)=
  0.95\frac{\mathcal L_P(u;a_k)}{\mathcal L_{P,\mathrm{pre}}}
  +0.05\frac{\mathcal L_D(u)}{\max(\rho^2,10^{-12})},
  \label{eq:grf-refine-merit}
\end{equation}
where $\mathcal L_{P,\mathrm{pre}}$ is the PDE loss at the constrained starting
pressure, lower-bounded by $10^{-12}$ and held fixed during the search.
If no step is accepted, that field keeps its previous $(u_k,a_k)$ for all
remaining rounds; other fields continue independently.  After an accepted
step, the Flow and decoder recompute $z$ and $a$ as shown in the algorithm.
We use $R=20$ rounds and report rounds 0 and 20.
For PINO, the inverse FNO reconstructs
$a_k=G_{\mathrm{PINO}}(u_k)$ at each round. Its own forward FNO supplies
$F_{\mathrm{PINO}}(a_0)$ to initialize the first pressure update; subsequent
updates start from the previously accepted $u_k$. This test-time pressure
correction uses the finite-volume residual in Eq.~\eqref{eq:grf-refine-losses},
while PINO training uses the CSRBF weak loss. All network weights remain
fixed, and no numerical Darcy solver is called inside the refinement loop.

\subsection{DGenNO results using the original optimization procedure}
\label{app:grf-dgenno-replay}
We recover continuous Darcy coefficients from coarse pressure observations
without clean anchors using our trained generators and DGenNO's original
Beta-MLP parameterization and optimization schedule~\citep{zang2025dgenno}. The test-time optimization
uses 1,000 Adam steps without multiple initializations or L-BFGS.
This supplements the direct-latent optimization used in the main comparison;
the two procedures are specified in Appendix~\ref{app:dgenno-inverse-protocol}.
The original DGenNO study uses a different continuous-coefficient family
and does not evaluate our $128\times128$ GRF Darcy setting.
Our models are trained for this setting and provide a reference comparison
(Table~\ref{tab:dgenno-replay-grf}).
\begin{table}[htbp]
\centering\scriptsize
\setlength{\tabcolsep}{3pt}
\caption{Continuous Darcy results using the original DGenNO optimization procedure on the coarse pressure grid ($N=64$ per setting). Errors are relative $L_2$.}
\label{tab:dgenno-replay-grf}
\begin{tabular}{@{}lccccc@{}}
\toprule
Method & $d$ & $\rho$ & Coefficient $a$ & Solver-$u$ & Time (s) \\
\midrule
DGenNO & 1024 & 0 & $0.71855\pm0.26673$ & $0.19793\pm0.08747$ & $24.01364\pm3.89045$ \\
DGenNO & 1024 & 0.01 & $0.66729\pm0.25667$ & $0.20109\pm0.10271$ & $24.22199\pm5.68484$ \\
DGenNO & 1024 & 0.03 & $0.71316\pm0.24072$ & $0.20244\pm0.08668$ & $25.00300\pm4.82461$ \\
\bottomrule
\end{tabular}
\end{table}

\subsection{forward prediction}
\label{app:grf-forward}
The encoder also predicts the pressure directly from
the permeability, $a\mapsto\hat u$.  This secondary capability is distinct
from solving the Darcy equation using a reconstructed $a$.  Results use the
same 64 held-out fields, with both the coefficient input
and pressure output evaluated on the full $128\times128$ grid.
The OCL-PDE/No-$Z$ row uses the retrained shared forward branch.
This branch has the lowest pressure relative $L_2$ error among the evaluated forward models.
\begin{center}
\vspace{-0.5\baselineskip}
\begingroup
\revisionsevencolor
\scriptsize
\setlength{\tabcolsep}{5pt}
\begin{tabular}{@{}lc@{}}
\toprule
Method & Direct $a\to u$ relative $L_2$ $\downarrow$ \\
\midrule
PINO & $0.003822$ \\
DGenNO ($d=1024$) & $0.01602$ \\
OCL-PDE/No-$Z$ shared branch & $0.002689$ \\
\bottomrule
\end{tabular}
\endgroup
\vspace{-0.5\baselineskip}
\end{center}

\subsection{Refinement evaluation with clean anchors}
\label{app:grf-refinement-evaluation}

\paragraph{Evaluation protocol.}
OCL-PDE, PINO, No-$Z$, and Conditional EDM+PDE use their
clean-anchor models throughout this study, with the same checkpoint reused
across anchor fractions and refinement rounds.
The study uses 16 held-out fields at $\rho\in\{1\%,3\%\}$.  The main table
uses 450 clean anchors among 900 interior nodes; the complete scan uses nested
fractions $0,5,\ldots,50\%$.  A seeded farthest-point
traversal~\citep{gonzalez1985clustering} defines a spatially covering anchor
order, with each fraction a prefix of the next.  Field, noise, Flow, and
anchor-order seeds are 49091, 59091, 69091, and 820812, respectively.
Clean anchor values are the sole explicit clean-pressure input.  The clean
coefficient truth, clean non-anchor pressures, and independent numerical-solver
pressure errors are used only for evaluation, never for the update.

\paragraph{Complete anchor sweep.}
We vary clean-anchor coverage from $0\%$ to $50\%$ of the 900 eligible
interior pressure locations and compare coefficient error before and after
refinement.
Before refinement, OCL-PDE is already competitive: coefficient error is
lowest at $1\%$ noise and close to No-$Z$ at $3\%$ noise.
Refinement starts to reduce OCL-PDE's error at $10\%$ anchor coverage;
its final error is lowest from $10\%$ coverage at $1\%$ noise and from
$15\%$ coverage at $3\%$ noise. With more anchors, the improvement grows,
reaching $34.9\%$ and $54.6\%$ relative error reductions at $50\%$ coverage.
At $0$--$5\%$ coverage, refinement instead increases OCL-PDE's error,
showing that sufficient clean information is needed for reliable correction
(Tables~\ref{tab:grf-anchor-sweep-rho1-low}--\ref{tab:grf-anchor-sweep-rho3-high}).

\begin{table}[H]
\centering
\scriptsize
\setlength{\tabcolsep}{3.2pt}
\caption{GRF coefficient relative $L_2$ at $\rho=1\%$ with 0--20\%
clean-anchor coverage. Each cell reports round 0 $\rightarrow$ round 20;
parentheses give the number of clean anchors. Lower is better.}
\label{tab:grf-anchor-sweep-rho1-low}
\resizebox{\textwidth}{!}{%
\begin{tabular}{@{}lccccc@{}}
\toprule
Method & \shortstack{0\%\\(0)} & \shortstack{5\%\\(45)} &
\shortstack{10\%\\(90)} & \shortstack{15\%\\(135)} &
\shortstack{20\%\\(180)} \\
\midrule
PINO & $0.04029\rightarrow0.04558$ & $0.04012\rightarrow0.04403$ & $0.04009\rightarrow0.04267$ & $0.03989\rightarrow0.04064$ & $0.03960\rightarrow0.03944$ \\
Conditional EDM+PDE & $0.10212\rightarrow0.11001$ & $0.10014\rightarrow0.10678$ & $0.09885\rightarrow0.10398$ & $0.09716\rightarrow0.10090$ & $0.09528\rightarrow0.09776$ \\
No-$Z$ ablation & $0.04013\rightarrow0.04370$ & $0.03992\rightarrow\mathbf{0.04182}$ & $0.03968\rightarrow0.04034$ & $0.03902\rightarrow0.03698$ & $0.03863\rightarrow0.03530$ \\
OCL-PDE & $0.03967\rightarrow\mathbf{0.04297}$ & $0.03934\rightarrow0.04217$ & $0.03910\rightarrow\mathbf{0.03884}$ & $0.03850\rightarrow\mathbf{0.03412}$ & $0.03798\rightarrow\mathbf{0.03027}$ \\
\bottomrule
\end{tabular}%
}
\end{table}

\begin{table}[H]
\centering
\scriptsize
\setlength{\tabcolsep}{3.2pt}
\caption{GRF coefficient relative $L_2$ at $\rho=1\%$ with 25--50\%
clean-anchor coverage. Each cell reports round 0 $\rightarrow$ round 20;
parentheses give the number of clean anchors. Lower is better.}
\label{tab:grf-anchor-sweep-rho1-high}
\resizebox{\textwidth}{!}{%
\begin{tabular}{@{}lcccccc@{}}
\toprule
Method & \shortstack{25\%\\(225)} & \shortstack{30\%\\(270)} &
\shortstack{35\%\\(315)} & \shortstack{40\%\\(360)} &
\shortstack{45\%\\(405)} & \shortstack{50\%\\(450)} \\
\midrule
PINO & $0.03946\rightarrow0.03874$ & $0.03936\rightarrow0.03784$ & $0.03914\rightarrow0.03729$ & $0.03892\rightarrow0.03705$ & $0.03875\rightarrow0.03686$ & $0.03836\rightarrow0.03669$ \\
Conditional EDM+PDE & $0.09441\rightarrow0.09654$ & $0.09468\rightarrow0.09653$ & $0.09486\rightarrow0.09633$ & $0.09525\rightarrow0.09651$ & $0.09647\rightarrow0.09754$ & $0.09808\rightarrow0.09912$ \\
No-$Z$ ablation & $0.03823\rightarrow0.03436$ & $0.03790\rightarrow0.03303$ & $0.03738\rightarrow0.03221$ & $0.03692\rightarrow0.03192$ & $0.03670\rightarrow0.03168$ & $0.03605\rightarrow0.03151$ \\
OCL-PDE & $0.03756\rightarrow\mathbf{0.02833}$ & $0.03727\rightarrow\mathbf{0.02549}$ & $0.03663\rightarrow\mathbf{0.02360}$ & $0.03581\rightarrow\mathbf{0.02324}$ & $0.03527\rightarrow\mathbf{0.02271}$ & $0.03456\rightarrow\mathbf{0.02249}$ \\
\bottomrule
\end{tabular}%
}
\end{table}

\begin{table}[H]
\centering
\scriptsize
\setlength{\tabcolsep}{3.2pt}
\caption{GRF coefficient relative $L_2$ at $\rho=3\%$ with 0--20\%
clean-anchor coverage. Each cell reports round 0 $\rightarrow$ round 20;
parentheses give the number of clean anchors. Lower is better.}
\label{tab:grf-anchor-sweep-rho3-low}
\resizebox{\textwidth}{!}{%
\begin{tabular}{@{}lccccc@{}}
\toprule
Method & \shortstack{0\%\\(0)} & \shortstack{5\%\\(45)} &
\shortstack{10\%\\(90)} & \shortstack{15\%\\(135)} &
\shortstack{20\%\\(180)} \\
\midrule
PINO & $0.06365\rightarrow0.06083$ & $0.06292\rightarrow\mathbf{0.05531}$ & $0.06223\rightarrow\mathbf{0.05116}$ & $0.06117\rightarrow0.04530$ & $0.05990\rightarrow0.04240$ \\
Conditional EDM+PDE & $0.10497\rightarrow0.12520$ & $0.10276\rightarrow0.11878$ & $0.10129\rightarrow0.11278$ & $0.09977\rightarrow0.10632$ & $0.09779\rightarrow0.10037$ \\
No-$Z$ ablation & $0.05806\rightarrow\mathbf{0.06051}$ & $0.05780\rightarrow0.05700$ & $0.05741\rightarrow0.05361$ & $0.05668\rightarrow0.04975$ & $0.05594\rightarrow0.04705$ \\
OCL-PDE & $0.06092\rightarrow0.06599$ & $0.06041\rightarrow0.06072$ & $0.05994\rightarrow0.05442$ & $0.05927\rightarrow\mathbf{0.04451}$ & $0.05827\rightarrow\mathbf{0.03895}$ \\
\bottomrule
\end{tabular}%
}
\end{table}

\begin{table}[H]
\centering
\scriptsize
\setlength{\tabcolsep}{3.2pt}
\caption{GRF coefficient relative $L_2$ at $\rho=3\%$ with 25--50\%
clean-anchor coverage. Each cell reports round 0 $\rightarrow$ round 20;
parentheses give the number of clean anchors. Lower is better.}
\label{tab:grf-anchor-sweep-rho3-high}
\resizebox{\textwidth}{!}{%
\begin{tabular}{@{}lcccccc@{}}
\toprule
Method & \shortstack{25\%\\(225)} & \shortstack{30\%\\(270)} &
\shortstack{35\%\\(315)} & \shortstack{40\%\\(360)} &
\shortstack{45\%\\(405)} & \shortstack{50\%\\(450)} \\
\midrule
PINO & $0.05858\rightarrow0.04053$ & $0.05782\rightarrow0.03887$ & $0.05667\rightarrow0.03791$ & $0.05548\rightarrow0.03750$ & $0.05395\rightarrow0.03727$ & $0.05252\rightarrow0.03694$ \\
Conditional EDM+PDE & $0.09668\rightarrow0.09789$ & $0.09680\rightarrow0.09735$ & $0.09676\rightarrow0.09642$ & $0.09698\rightarrow0.09636$ & $0.09812\rightarrow0.09739$ & $0.09927\rightarrow0.09893$ \\
No-$Z$ ablation & $0.05516\rightarrow0.04560$ & $0.05487\rightarrow0.04344$ & $0.05411\rightarrow0.04221$ & $0.05316\rightarrow0.04156$ & $0.05204\rightarrow0.04085$ & $0.05127\rightarrow0.04034$ \\
OCL-PDE & $0.05731\rightarrow\mathbf{0.03597}$ & $0.05691\rightarrow\mathbf{0.03019}$ & $0.05625\rightarrow\mathbf{0.02633}$ & $0.05488\rightarrow\mathbf{0.02566}$ & $0.05317\rightarrow\mathbf{0.02457}$ & $0.05239\rightarrow\mathbf{0.02381}$ \\
\bottomrule
\end{tabular}%
}
\end{table}

\subsection{Robustness to noisy coarse pressure observations}
\label{app:grf-noise-robustness}
We assess continuous-coefficient Darcy recovery when all interior
coarse-grid pressure observations are subject to additive Gaussian noise,
without clean anchors or test-time pressure refinement. On the aligned
$32\times32$ grid, the 900 interior values satisfy
$y_i=u_i+\epsilon_i$, where $\epsilon_i\sim\mathcal N(0,(\rho s_u)^2)$
and $s_u$ is estimated from the training set. We evaluate
$\rho\in\{0,0.01,0.03\}$, with $\rho=0$ providing the clean control;
the known zero boundary remains exact. Results are averaged over 64 held-out
fields. This experiment complements the clean-anchor study in the main text
by measuring sensitivity to noise when no higher-quality measurements are available.
OCL-PDE, PINO, No-$Z$, and Conditional EDM+PDE use the same checkpoints
as in the clean-anchor evaluation, with zero clean anchors and no pressure refinement. OCL-PDE uses $K=64$ samples and 50 Euler
steps; Conditional EDM+PDE uses its final raw weights, $K=64$ samples, and
25 Karras--Heun steps. Both average samples in physical coefficient space.
Solver-$u$ is evaluated on all 64 fields at each noise level.

\begin{table}[H]
\centering
\scriptsize
\setlength{\tabcolsep}{1.4pt}
\caption{Robustness of continuous Darcy recovery to additive pressure noise on the aligned $32\times32$
pressure grid. Independent Gaussian noise with standard
deviation $\rho s_u$ is added to all interior pressure values;
the boundary remains zero.
The case $\rho=0$ is noise-free. Reported coefficient and solver-$u$ errors are means over 64 held-out fields.
For each method with recorded timing, test time is the mean
per-field inference time over all three noise levels, excluding numerical
Darcy solves; ``--'' denotes unavailable timing. Solver-$u$ denotes an independent numerical Darcy solve from each reconstructed coefficient.  Lower is better. }

\label{tab:grf-low-fidelity}
\begin{tabular}{@{}lccccccc@{}}
\toprule
 {Method} &
\multicolumn{3}{c}{$a$ relative $L_2$} &
\multicolumn{3}{c}{Solver-$u$ relative $L_2$} & \shortstack{Mean time\\(s/field)} \\
\cmidrule(lr){2-4}\cmidrule(lr){5-7}\cmidrule(l){8-8}
&  {$\rho=0$} &  {$\rho=0.01$} &  {$\rho=0.03$}
&  {$\rho=0$} &  {$\rho=0.01$} &  {$\rho=0.03$} & \\
\midrule
PINO & $0.03561$ & $0.03986$ & $0.06239$ & $0.00430$ & $0.00471$ & $0.00711$ & $0.00141$ \\
{DGenNO ($d=1024$)} & {$0.15056$} & {$0.14958$} & {$0.15234$} & {$0.03358$} & {$0.03316$} & {$0.03405$} & $99.74$ \\
Conditional EDM+PDE & $0.09575$ & $0.09636$ & $0.09943$ & $0.02633$ & $0.02637$ & $0.02674$ & $12.25$ \\
No-$Z$ ablation & $0.02742$ & $0.03858$ & $\mathbf{0.05704}$ & $0.00318$ & $0.00371$ & $0.00629$ & $0.01571$ \\
OCL-PDE & $\mathbf{0.01807}$ & $\mathbf{0.03686}$ & $0.05740$ & $\mathbf{0.00192}$ & $\mathbf{0.00334}$ & $\mathbf{0.00620}$ & $2.541$ \\
\bottomrule
\end{tabular}
\end{table}

Pressure error is lowest for OCL-PDE at all three noise levels, while
coefficient error is lowest at $\rho=0$ and $1\%$
(Table~\ref{tab:grf-low-fidelity}). At $\rho=3\%$, No-$Z$ has a slightly lower
coefficient error ($0.05704$ versus $0.05740$).
OCL-PDE's coefficient error rises from $0.01807$ without noise to $0.05740$
at $3\%$ noise, but remains below all external baselines at every tested
noise level. Pressure error increases from $0.00192$ to $0.00620$ while
retaining the lowest error. Thus, reconstruction degrades with noise but
remains controlled over the tested range without clean anchors or refinement.

\section{Binary-Darcy implementation and additional results}
\label{app:binary-implementation}
\subsection{Latent construction}
\label{app:binary-latent}
The latent branch directly encodes the complete phase geometry.
Let $S(a)$ be the signed Euclidean distance to the phase boundary, positive
in the high-permeability phase, divided by eight pixels and clipped to
$[-1,1]$~\citep{curless1996volumetric}. We use the full-field input $h=S(a)/2$
and apply zero-input centering to the latent mean:
\begin{equation}
  h=\frac{S(a)}{2},\qquad
  \mu_z=E^z_{\theta_E}(h)-E^z_{\theta_E}(0).
  \label{eq:binary-latent-design}
\end{equation}
The latent has shape $4\times16\times16$. The signed-distance
transform is used only by the latent encoder; the forward branch takes the
physical coefficient, and the decoder predicts phase logits.

\subsection{Data, networks, and training}
\label{app:binary-data-training}

\paragraph{Observation sampling.}
During training, the random-point observation rate is sampled over the full
$2\%$--$10\%$ range, rather than restricted to the three evaluation rates;
$2\%$, $5\%$, and $10\%$ are used only for formal evaluation and reporting.

\paragraph{Network inputs and reconstruction.}
Binary Darcy retains the Darcy FNO/CNN/U-Net-style core.
The response operator receives the physical coefficient
$a\in\{3,12\}^{B\times1\times128\times128}$, coordinates, and forcing, and
returns the full pressure field. The decoder receives $z$, raw zero-filled
pressure $y_q=M_q\odot u\in\mathbb R^{B\times1\times128\times128}$, its exact
binary mask $M_q$, coordinates, and forcing. Observations are random interior
points with no interpolation. The fixed test masks at $2\%$, $5\%$, and
$10\%$ are nested and contain 318, 794, and 1588 points, respectively.

The decoder combines an observation-conditioned base branch
with a latent correction to predict binary logits, and is trained with
binary cross-entropy with logits. A sigmoid gives the high-phase probability; we
average it over $K=16$ samples and threshold the mean at $0.5$.
\paragraph{Flow conditioning.}
The decoder receives pressure in physical units. The Flow
divides its pressure input once by the training-set standard deviation
$s_u$ and receives
\begin{equation}
  \eta_{\mathrm{bin}}=[y_q/s_u,M_q]\quad\text{on }128^2,
  \label{eq:binary-flow-condition}
\end{equation}
The Flow is conditioned on observed pressure and its mask.
Its FPN-inspired condition extractor maps these inputs to the
$16\times16$ latent grid. The velocity backbone uses width 192, four spatially
modulated residual blocks, one four-head self-attention block, and four
further residual blocks.

\paragraph{Training and evaluation settings.}
The AE has 30,313,272 parameters; the 20-parameter log-variance
head and 10,503,668-parameter Flow give a total of 40,816,960.
Training uses a frozen physical forward branch and 1,000
paired training fields. The base is trained for 100 epochs with 100 updates
per epoch and batch size 8, using paired numerical pressure observations
and a $5\times$ binary cross-entropy loss. With the base and forward branch
frozen, the full-$z$ encoder and decoder's latent path are trained for
100 epochs with 100 updates per epoch and batch size 4, using paired and
generated fields with reconstruction and PDE losses. The autoencoder is
then frozen for 100 epochs of pressure-and-mask-conditioned Flow training,
with 100 updates per epoch, batch size 8, four observation views, and equal
probabilities of paired numerical and dynamically predicted pressures.

Evaluation uses the same 64 held-out fields for all methods;
OCL-PDE generates $K=16$ samples with 50 Euler steps. PINO uses epoch-110
weights, DiffusionPDE the epoch-350 EMA checkpoint, and DGenNO its final
$d=1024$ weights.

\subsection{Additional reconstruction metrics}
\label{app:binary-results}
We assess interface recovery using Boundary F1, ASSD, and HD95 on the same
64 held-out fields and three sparse observation masks as in the main experiment.
OCL-PDE achieves the highest Boundary F1 and the lowest mean ASSD and HD95
at all three observation rates (Table~\ref{tab:binary-random-boundary}).

\begin{table}[htbp]
\centering
\scriptsize
\setlength{\tabcolsep}{2.4pt}
\caption{Additional Binary Darcy metrics at $2\%$, $5\%$, and
$10\%$ observation rates. Boundary F1 uses a one-pixel tolerance; ASSD and
HD95 are measured in pixels. Bold values are best within each observation rate.}
\label{tab:binary-random-boundary}
\begin{tabular*}{\linewidth}{@{\extracolsep{\fill}}l*{9}{r}@{}}
\toprule
Method & \multicolumn{3}{c}{2\% (318 points)} & \multicolumn{3}{c}{5\% (794 points)} & \multicolumn{3}{c}{10\% (1588 points)} \\
\cmidrule(lr){2-4}\cmidrule(lr){5-7}\cmidrule(l){8-10}
 & \shortstack{Boundary\\F1 $\uparrow$} & \shortstack{ASSD\\(px) $\downarrow$} & \shortstack{HD95\\(px) $\downarrow$} & \shortstack{Boundary\\F1 $\uparrow$} & \shortstack{ASSD\\(px) $\downarrow$} & \shortstack{HD95\\(px) $\downarrow$} & \shortstack{Boundary\\F1 $\uparrow$} & \shortstack{ASSD\\(px) $\downarrow$} & \shortstack{HD95\\(px) $\downarrow$} \\
\midrule
PINO & $0.47774$ & $3.43339$ & $13.20167$ & $0.60188$ & $2.40210$ & $9.76224$ & $0.65318$ & $2.12762$ & $9.99703$ \\
DGenNO ($d=1024$) & $0.49796$ & $5.12891$ & $21.74188$ & $0.50092$ & $5.21081$ & $22.72245$ & $0.49330$ & $5.39692$ & $22.19505$ \\
DiffusionPDE & $0.3540$ & $7.6068$ & $29.0139$ & $0.4270$ & $6.5002$ & $26.5313$ & $0.4493$ & $6.1458$ & $25.1457$ \\
No-$Z$ & $0.6279$ & $2.8237$ & $13.3906$ & $0.7514$ & $2.0703$ & $10.8046$ & $0.7894$ & $1.8443$ & $9.4692$ \\
OCL-PDE & $\mathbf{0.7509}$ & $\mathbf{1.9840}$ & $\mathbf{10.0763}$ & $\mathbf{0.8721}$ & $\mathbf{1.2133}$ & $\mathbf{6.4314}$ & $\mathbf{0.8943}$ & $\mathbf{0.9803}$ & $\mathbf{5.2465}$ \\
\bottomrule
\end{tabular*}
\end{table}

The median field-wise HD95 values for OCL-PDE are $5.8310$, $4.0246$, and $3.0000$ pixels at $2\%$, $5\%$, and $10\%$ observation rates, respectively,
compared with $9.21954$, $6.32456$, and $5.38516$ pixels for PINO.
Its mean IoU values are $0.93492$, $0.95610$, and $0.96061$,
and mean Dice scores are $0.96615$, $0.97744$, and $0.97982$, respectively.

\subsection{DGenNO with the original inference procedure}
\label{app:binary-dgenno-replay}
We apply DGenNO's original inverse inference procedure to binary coefficient
recovery from random pressure observations. The original Darcy experiments
use a $29\times29$ grid~\citep{zang2025dgenno}; we train the adapted model
on our $128\times128$ data to provide a reference comparison
(Table~\ref{tab:dgenno-replay-binary}).
\begin{table}[htbp]
\centering\scriptsize
\setlength{\tabcolsep}{3pt}
\caption{DGenNO with the original inference procedure for Binary Darcy from random pressure observations ($N=64$ per setting). Entries are mean $\pm$ sample standard deviation. Both errors are reported as ratios.}
\label{tab:dgenno-replay-binary}
\begin{tabular}{@{}lccccc@{}}
\toprule
Method & $d$ & Observed & Hard error & Solver-$u$ & Time (s) \\
\midrule
DGenNO & 1024 & $2\%$ & $0.16463\pm0.07715$ & $0.13248\pm0.06821$ & $25.21020\pm5.67132$ \\
DGenNO & 1024 & $5\%$ & $0.18134\pm0.10124$ & $0.14612\pm0.08035$ & $26.16990\pm2.63666$ \\
DGenNO & 1024 & $10\%$ & $0.17399\pm0.07537$ & $0.15497\pm0.09822$ & $24.58750\pm5.65273$ \\
\bottomrule
\end{tabular}
\end{table}

\subsection{forward prediction}
\label{app:binary-forward}
We evaluate the same 64 held-out coefficient fields on the full
$128\times128$ pressure grid.
DiffusionPDE uses the epoch-350 EMA joint $(a,u)$ model to predict
pressure conditioned on the full ground-truth coefficient field, without
pressure observations, using one 2,000-step sample per field. We compare
the generated pressure with the ground-truth pressure.
OCL-PDE achieves the lowest pressure relative $L_2$ error among the evaluated methods.
\begin{center}
\vspace{-0.5\baselineskip}
\begingroup
\revisionsevencolor
\scriptsize
\setlength{\tabcolsep}{5pt}
\begin{tabular}{@{}lc@{}}
\toprule
Method & Pressure relative $L_2$ $\downarrow$ \\
\midrule
PINO & $0.01447$ \\
DGenNO ($d=1024$) & $0.05679$ \\
DiffusionPDE (conditional joint sampling) & $0.154958$ \\
No-$Z$ (matched) & $0.028866$ \\
OCL-PDE & $0.012852$ \\
\bottomrule
\end{tabular}
\endgroup
\vspace{-0.5\baselineskip}
\end{center}

\subsection{Robustness and diagnostic studies}
\label{app:robustness-diagnostics}

\paragraph{Binary-Darcy direct noise stress test.}
We compare the clean-trained OCL-PDE, No-$Z$, and PINO models from the
main experiment without retraining or test-time refinement. All methods
use the same 64 held-out fields, nested pressure masks at $2\%$, $5\%$,
and $10\%$ observation density, and identical noisy pressure inputs.
OCL-PDE uses $K=16$ samples and 50 Euler steps. Gaussian noise is added
only at observed interior points; unobserved locations remain zero,
with no interpolation or noise-level token.
For the reported target PSNR levels $P\in\{80,50\}$ dB, we set
$\sigma=R_u10^{-P/20}$, where the global pressure range
$R_u\approx0.01742$ is estimated from training data only. The corresponding
noise standard deviations are $1.741593\times10^{-6}$ and
$5.507402\times10^{-5}$, respectively.
Each field uses paired OCL-PDE sampling seeds and a shared standard-normal
noise grid across densities and noise levels. The clean No-$Z$ and PINO
probability arrays exactly reproduce the saved main-experiment outputs.

\begin{table}[ht]
\centering
\scriptsize
\setlength{\tabcolsep}{7pt}
\caption{Binary Darcy reconstruction under observation noise without
retraining or refinement. Mean hard pixel errors over 64 fields are
reported as fractions.}
\label{tab:binary-noise-comparison}
\begin{tabular}{@{}llrrr@{}}
\toprule
Observation density & Method & Clean & 80 dB & 50 dB \\
\midrule
$2\%$ & PINO & $0.068003$ & $0.067999$ & $0.068479$ \\
 & No-$Z$ & $0.046590$ & $0.046590$ & $0.047270$ \\
 & OCL-PDE & $\mathbf{0.035237}$ & $\mathbf{0.035226}$ & $\mathbf{0.036256}$ \\
\midrule
$5\%$ & PINO & $0.050697$ & $0.050689$ & $0.050801$ \\
 & No-$Z$ & $0.034540$ & $0.034532$ & $0.035040$ \\
 & OCL-PDE & $\mathbf{0.023565}$ & $\mathbf{0.023566}$ & $\mathbf{0.024655}$ \\
\midrule
$10\%$ & PINO & $0.045720$ & $0.045725$ & $0.045819$ \\
 & No-$Z$ & $0.031425$ & $0.031434$ & $0.031753$ \\
 & OCL-PDE & $\mathbf{0.021048}$ & $\mathbf{0.021038}$ & $\mathbf{0.021707}$ \\
\bottomrule
\end{tabular}
\end{table}
Hard pixel error changes little at the reported 80 and 50 dB noise levels.
OCL-PDE has lower error than both No-$Z$ and PINO at every observation
density under these two noise levels (Table~\ref{tab:binary-noise-comparison}).
At 50 dB, its absolute error increase is $0.001019$, $0.001090$, and
$0.000659$ at $2\%$, $5\%$, and $10\%$ density, respectively.

\paragraph{Binary-Darcy DiffusionPDE sanity check.}
Our DiffusionPDE model is trained on only 1,000 supervised field pairs.
Its weaker pressure-only recovery motivates a check of whether the learned
joint model can use additional information; this limited-data evaluation
differs from the original DiffusionPDE benchmark~\citep{huang2024diffusionpde}.
We use the same epoch 350 EMA checkpoint as in the Binary
Darcy main experiment to examine the effect of additional coefficient
observations. We retain the held-out field and random pressure masks used
in the main comparison (Figure~\ref{fig:binary-random-main}).
At each sparsity, we compare pressure-only
recovery with recovery using 500 additional coefficient values at fixed
random locations (mask seed 1). Both runs use 2,000 sampling steps and the
same initial sampling noise.

\begin{figure}[ht]
  \centering
  \includegraphics[width=0.68\textwidth]{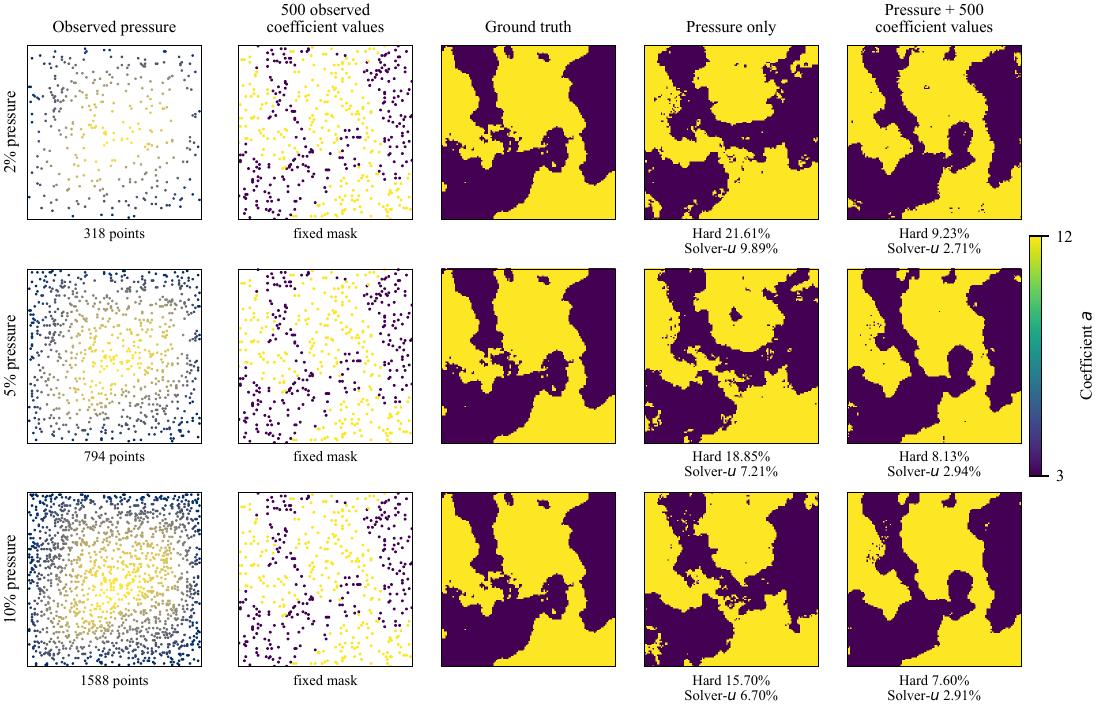}
  \caption{DiffusionPDE coefficient-observation sanity check
  with the main experiment's epoch 350 EMA checkpoint. Rows show the same
  held-out field at $2\%$, $5\%$, and $10\%$ random pressure
  observation rates. Columns show observed pressure, the 500 fixed
  coefficient observations, ground truth, pressure-only recovery, and
  recovery with both types of observations. Reconstructed coefficients are
  thresholded at the phase midpoint; annotations report hard pixel error
  and solver-$u$ relative $L_2$ error from an independent numerical solve.}
  \label{fig:binary-diffusionpde-a500-sanity}
\end{figure}

Adding coefficient observations reduces both hard pixel error
and solver-$u$ error at all three sparsities for this field
(Figure~\ref{fig:binary-diffusionpde-a500-sanity}). The substantial improvement shows that the model responds to additional
coefficient information, providing evidence against a collapsed joint prior.
This check supports the functioning of the trained model, but does not
by itself rule out all training or implementation issues.

\clearpage
\begingroup
\raggedbottom
\setlength{\floatsep}{10pt plus 2pt minus 2pt}
\setlength{\textfloatsep}{10pt plus 2pt minus 2pt}
\setlength{\intextsep}{10pt plus 2pt minus 2pt}
\makeatletter
\setlength{\@fptop}{0pt}
\setlength{\@fpsep}{12pt}
\setlength{\@fpbot}{0pt plus 1fil}
\makeatother
\section{Supplementary ablation results}
\label{app:supplementary-results}

\subsection{Complementary-latent interventions}
\label{app:latent-evidence}
\revisionseven{Across the four directed reconstructions from two fixed pairs,
the mean matched relative $L_2$ error is $0.0261$.  It increases to $0.7233$
when $U$ is swapped, $0.7862$ when $Z$ is swapped, $0.9968$ when $Z$ is set to
zero, and $0.5527$ when $U$ is set to zero
(Figure~\ref{fig:ac-zu-cross-full}). These two-pair results are mechanism
evidence rather than population statistics.}

\begin{figure}[!ht]
  \centering
  \includegraphics[width=0.7\textwidth]{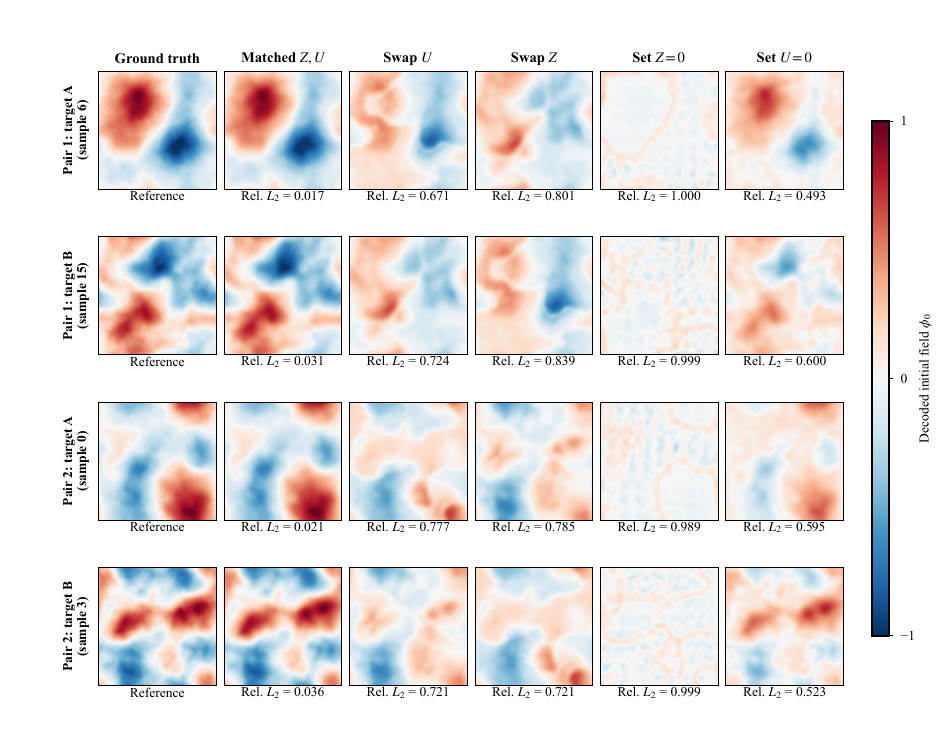}
  \caption{\revisionseven{Full Allen--Cahn decoder-branch intervention for two
  fixed test pairs: trajectories 6 and 15 in the first two rows, and
  trajectories 0 and 3 in the last two rows.  Columns show the reference field, matched reconstruction, exchanged observation,
  exchanged latent, zero latent, and zero observation. The plot labels $U$ and $Z$
  denote observation $y$ and latent $z$, respectively. Each reconstruction is annotated with its relative $L_2$ error.}}
  \label{fig:ac-zu-cross-full}
\end{figure}

\subsection{What the latent adds to a fixed-observation reconstruction}
\label{app:latent-corrections}

\paragraph{Protocol and scope.}
We perform inference-only diagnostics with one trained checkpoint pair
per task (training seed 0), using all 20 held-out Allen--Cahn trajectories
and all 64 held-out GRF Darcy fields. Networks are frozen and evaluated
with identical observations and decoder conditions across interventions.
Allen--Cahn uses the clean full terminal field at $t=4$ and reuses the
saved formal Flow posterior mean (inference seed 9091). GRF uses its mixed-observation AE/Flow checkpoints at $\rho=0$ with the native $32\times32$
pressure grid (900 interior measurements), no clean anchors, and no
pressure refinement. Its inference seed is 69091. Both generated estimates
use $K=64$ samples and 50 Euler steps. The GRF diagnostic uses $K=64$, compared with $K=16$ in
the clean-anchor experiment. Test IDs 0 and 1, the first two in canonical stored order, were
fixed for visualization before computing errors.

Binary Darcy uses the AE and Flow from the main experiment, evaluated on the same 64 fields at $2\%$, $5\%$,
and $10\%$ observation rates. Inference uses $K=16$, 50 Euler steps, and fixed per-field Gaussian seeds $29091+i$.
Its generated probabilities exactly reproduce the formal test outputs.
All three reconstructions share the same AE and observations; figures use
the first two stored fields, IDs 0 and 1, at $5\%$ observations.

\paragraph{Three reconstructions from the same decoder.}
Let $f=\phi(\cdot,0)$ for Allen--Cahn and $f=g=\log a$ for GRF Darcy.
For Binary Darcy, $f=\log a$ is computed from the hard
coefficient field $a\in\{3,12\}$.
Write $D_{\theta_D}(z,y)$ for the decoder output in this analysis field, including
the $\tanh$ postprocessing for GRF and the identity transform for
Allen--Cahn. For Binary Darcy, the decoder logits are
converted to probabilities and thresholded at $0.5$ before taking
$\log a$. We compare
\begin{equation}
  f_0=D_{\theta_D}(z=0,y),\qquad
  f_{\mathrm{orc}}=D_{\theta_D}(z^\star,y),\qquad
  r^\star=f_{\mathrm{true}}-f_0,\qquad
  \Delta f_m=\tilde f_m-f_0.
  \label{eq:latent-correction-fields}
\end{equation}
Here $z^\star=\mu$ is the encoder mean obtained from the
ground-truth field, and $r^\star$ is the correction needed beyond $f_0$.
The oracle tests the representation and decoder, uses ground truth, and is
not an available inverse estimator or a guaranteed error lower bound.
The $D_{\theta_D}(z=0,y)$ output uses the same checkpoint as the other two
reconstructions and is distinct from the separately trained,
parameter-matched No-$Z$ baseline. For Allen--Cahn, literal $z=0$ does not
exactly disable the latent branch because its decoder lacks the exact
zero-correction centering used in GRF; its error should not be interpreted
as the amount of information recoverable from the observation alone.

The Flow estimate uses only the observation and observation-derived
conditions. Allen--Cahn averages decoded initial fields. GRF averages in
physical coefficient space:
\begin{equation}
  \widehat a_{\mathrm{gen}}=\frac{1}{K}\sum_{k=1}^{K}
      \exp G(z^{(k)},y),\qquad
  \widehat g_{\mathrm{gen}}=\log\widehat a_{\mathrm{gen}}.
  \label{eq:latent-diagnostic-mean-a}
\end{equation}
Latents are mapped back from standardized Flow coordinates to encoder
coordinates before decoding. Neither a mean latent nor a mean log-field
is substituted for Eq.~\eqref{eq:latent-diagnostic-mean-a}.
GRF uses the same native-grid Flow checkpoint as the main
experiments, without a base-field condition.
For Binary Darcy, the Flow point estimate averages the
$K=16$ sigmoid probabilities before thresholding at $0.5$ and mapping to
$a\in\{3,12\}$. The diagnostic analysis then uses its logarithm.

\paragraph{Metrics.}
We report relative $L_2$ error in the physical field
($\phi(\cdot,0)$ or $a$), and in the analysis field $f$.
For a detail-sensitive diagnostic, $P_{\mathrm{hi}}$ selects coefficients
of an orthonormal two-dimensional DCT-II with zero-based indices satisfying
$k_x^2+k_y^2\ge16^2$. We compute
\begin{equation}
  E_f=\frac{\|\tilde f-f_{\mathrm{true}}\|_2}
               {\|f_{\mathrm{true}}\|_2},\qquad
  E_{\mathrm{hi}}=\frac{\|P_{\mathrm{hi}}(\tilde f-f_{\mathrm{true}})\|_2}
                       {\|P_{\mathrm{hi}}f_{\mathrm{true}}\|_2},\qquad
  Q=1-\frac{\sum_i\|\tilde f_i-f_{\mathrm{true},i}\|_2^2}
            {\sum_i\|f_{0,i}-f_{\mathrm{true},i}\|_2^2}.
  \label{eq:latent-correction-metrics}
\end{equation}
Thus $Q$ pools squared errors before taking the ratio, and is not clipped.
Metrics are accumulated in float64 with a common denominator floor of
$10^{-12}$; no test field activates this floor. Standard deviations are
sample standard deviations across held-out fields, not across training
seeds. High-band error measures recovery of the correct spatial details,
rather than merely the presence of high-frequency energy.

\begin{table}[!htbp]
\centering
\caption{Fixed-observation latent diagnostics. Errors are
percentages (mean $\pm$ sample standard deviation); $Q$ is the pooled
analysis-field squared-error reduction relative to $z=0$.
The oracle uses the ground-truth encoder mean $\mu$.
Flow uses $K=64$ samples for Allen--Cahn and GRF Darcy, and $K=16$ for
Binary Darcy. Binary outputs are thresholded after averaging phase
probabilities; hard pixel error applies only to Binary Darcy.}
\label{tab:latent-correction-metrics}
\scriptsize
\setlength{\tabcolsep}{3pt}
\begin{tabular}{@{}lccccc@{}}
\toprule
Reconstruction & Hard error & $E_{\mathrm{physical}}$ & $E_f$ & $E_{\mathrm{hi}}$ & $Q$ \\
\midrule
\multicolumn{6}{l}{\textit{Allen--Cahn: $n=20$, $f=\phi(\cdot,0)$}} \\
$D_{\theta_D}(z=0,y)$ & -- & $100.67\pm0.62$ & $100.67\pm0.62$ & $188.76\pm23.43$ & $0$ \\
$D_{\theta_D}(z^\star,y)$ & -- & $2.66\pm0.72$ & $2.66\pm0.72$ & $55.62\pm3.47$ & $0.9993$ \\
$D_{\theta_D}(z_{flow},y)$ & -- & $5.90\pm1.82$ & $5.90\pm1.82$ & $78.95\pm3.99$ & $0.9964$ \\
\midrule
\multicolumn{6}{l}{\textit{GRF Darcy: $n=64$, $f=\log a$}} \\
$D_{\theta_D}(z=0,y)$ & -- & $3.66\pm0.93$ & $9.08\pm2.51$ & $97.98\pm2.42$ & $0$ \\
$D_{\theta_D}(z^\star,y)$ & -- & $0.57\pm0.16$ & $1.40\pm0.45$ & $32.11\pm2.19$ & $0.9762$ \\
$D_{\theta_D}(z_{flow},y)$ & -- & $1.81\pm0.50$ & $4.48\pm1.28$ & $83.80\pm5.10$ & $0.7553$ \\
\midrule
\multicolumn{6}{l}{\textit{Binary Darcy: $n=64$, $f=\log a$, $2\%$ observations}} \\
$D_{\theta_D}(z=0,y)$ & $5.54\pm2.56$ & $23.23\pm5.33$ & $16.33\pm3.77$ & $121.94\pm6.16$ & $0.0000$ \\
$D_{\theta_D}(z^\star,y)$ & $0.78\pm0.54$ & $8.26\pm3.22$ & $5.81\pm2.28$ & $49.28\pm12.62$ & $0.8600$ \\
$D_{\theta_D}(z_{flow},y)$ & $3.52\pm1.66$ & $18.50\pm4.29$ & $13.01\pm3.05$ & $110.39\pm5.63$ & $0.3637$ \\
\midrule
\multicolumn{6}{l}{\textit{Binary Darcy: $n=64$, $f=\log a$, $5\%$ observations}} \\
$D_{\theta_D}(z=0,y)$ & $4.48\pm2.05$ & $20.87\pm4.82$ & $14.68\pm3.41$ & $117.02\pm6.18$ & $0.0000$ \\
$D_{\theta_D}(z^\star,y)$ & $0.66\pm0.54$ & $7.46\pm3.32$ & $5.26\pm2.36$ & $46.05\pm13.71$ & $0.8519$ \\
$D_{\theta_D}(z_{flow},y)$ & $2.36\pm1.19$ & $15.03\pm3.82$ & $10.57\pm2.71$ & $96.43\pm6.59$ & $0.4739$ \\
\midrule
\multicolumn{6}{l}{\textit{Binary Darcy: $n=64$, $f=\log a$, $10\%$ observations}} \\
$D_{\theta_D}(z=0,y)$ & $4.34\pm2.00$ & $20.50\pm4.76$ & $14.42\pm3.39$ & $116.27\pm5.99$ & $0.0000$ \\
$D_{\theta_D}(z^\star,y)$ & $0.65\pm0.48$ & $7.51\pm3.10$ & $5.29\pm2.20$ & $47.09\pm13.41$ & $0.8491$ \\
$D_{\theta_D}(z_{flow},y)$ & $2.10\pm1.07$ & $14.22\pm3.59$ & $10.00\pm2.55$ & $92.54\pm6.70$ & $0.5148$ \\
\bottomrule
\end{tabular}
\end{table}

\paragraph{Results.}
The latent supplies different corrections across tasks. In GRF Darcy, Flow approximately
halves coefficient error ($3.66\%\rightarrow1.81\%$) and removes $75.5\%$
of the pooled log-coefficient squared error ($Q=0.7553$). The accompanying
decrease in high-frequency error ($97.98\%\rightarrow83.80\%$) shows that
the latent correction also improves the spatial details measured by this
spectral diagnostic (Table~\ref{tab:latent-correction-metrics}).

For Binary Darcy, Flow corrects phase assignments: at $5\%$
observations, hard pixel error falls from $4.48\%$ to $2.36\%$, approximately
halving the misclassified area. Across the three observation densities,
it removes $36.4$--$51.5\%$ of the pooled log-coefficient squared error.
For Allen--Cahn, Flow reduces initial-field error to $5.90\%$ and removes
$99.64\%$ of the zero-latent squared error. Here the latent contributes to
the broad initial-field structure as well as finer corrections; the large
gain is relative to a zero-latent path that was not separately trained
for reconstruction.

The oracle has lower field and high-frequency errors in
every setting. For example, GRF high-frequency error falls to $32.11\%$
with the oracle, compared with $83.80\%$ for Flow. Thus the decoder can
represent finer reconstructions than those recovered through
observation-conditioned sampling. The spatial corrections also recover
part of the structure missing from the zero-latent outputs in the two
illustrated fields per task
(Figures~\ref{fig:latent-correction-grf}--\ref{fig:latent-correction-binary}).

\begin{figure}[!htbp]
\centering
\includegraphics[width=0.8\textwidth]{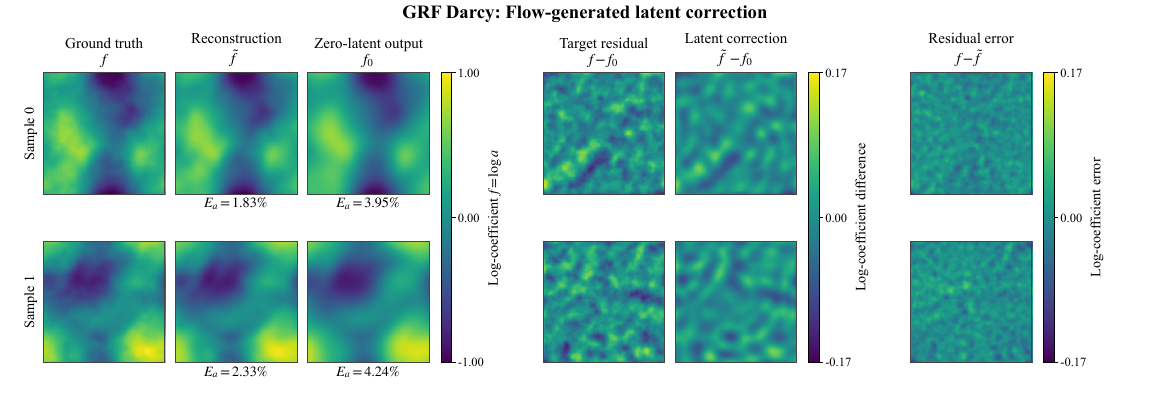}
\caption{\textbf{GRF Darcy latent corrections.}
Flow-generated reconstruction and zero-latent base, with the target residual,
latent correction, and remaining error in $f=\log a$.
We report coefficient reconstruction errors in relative $L_2$.}
\label{fig:latent-correction-grf}
\end{figure}

\begin{figure}[!htbp]
\centering
\includegraphics[width=0.8\textwidth]{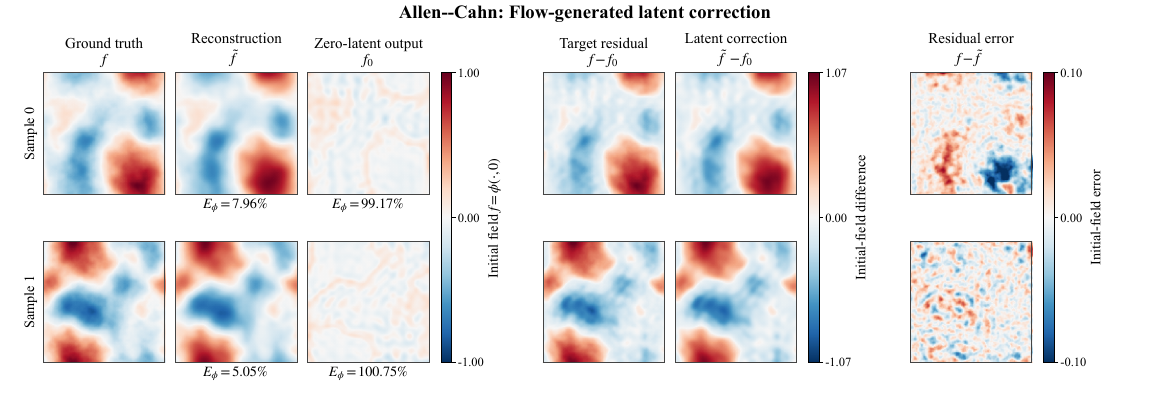}
\caption{\textbf{Allen--Cahn latent corrections.}
Flow-generated reconstruction and zero-latent base, with the target residual,
latent correction, and remaining error in $f=\phi(\cdot,0)$.
We report initial-field reconstruction errors in relative $L_2$.}
\label{fig:latent-correction-ac}
\end{figure}

\begin{figure}[!htbp]
\centering
\includegraphics[width=0.8\textwidth]{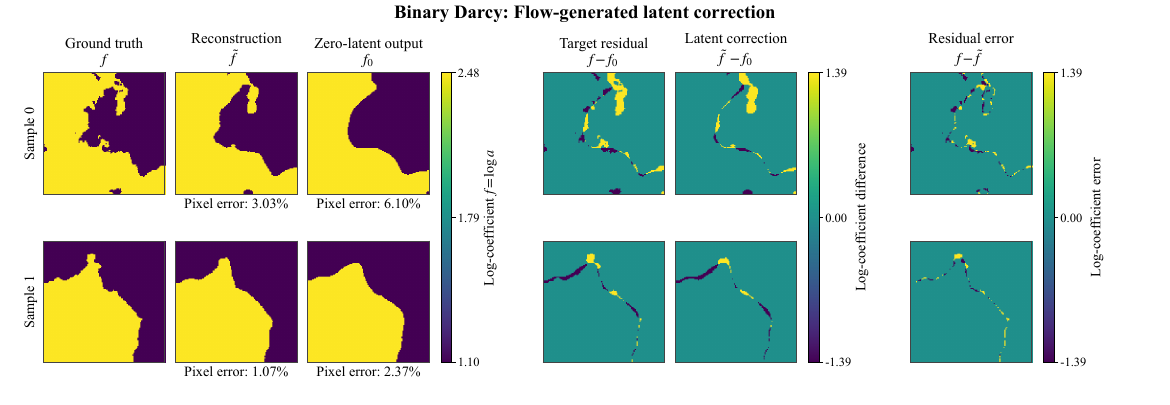}
\caption{\textbf{Binary Darcy latent corrections at $5\%$ observations.}
Flow-generated reconstruction and zero-latent base, with the target residual,
latent correction, and remaining error in $f=\log a$.
We report hard pixel errors.}
\label{fig:latent-correction-binary}
\end{figure}

\clearpage
\endgroup

\end{document}